\documentclass[lettersize,journal]{IEEEtran}
\usepackage{amsmath,amsfonts}
\usepackage{algorithmic}
\usepackage{algorithm}
\usepackage{array}
\usepackage[caption=false,font=normalsize,labelfont=sf,textfont=sf]{subfig}
\usepackage{textcomp}
\usepackage{stfloats}
\usepackage{url}
\usepackage{verbatim}
\usepackage{graphicx}
\usepackage{subfig}
 \usepackage{multirow}
\usepackage{booktabs}
\usepackage[justification=centering]{caption}

\begin{document}

\title{Multi-View Semi-Supervised Label Distribution Learning with Local Structure Complementarity}

\author{Yanshan Xiao,
        Kaihong Wu
         and Bo Liu$^*$
\thanks{Yanshan Xiao and Kaihong Wu are with the School of
Computers, Guangdong University of Technology.}
\thanks{Bo Liu (Corresponding Author) is with the School of Automation, Guangdong University of Technology.}  
\thanks{This work has been submitted to the IEEE for possible publication. Copyright may be transferred without notice, after which this version may no longer be accessible.}
 \vspace{-0.5cm}
  } 



\maketitle

\begin{abstract}
Label distribution learning (LDL) is a paradigm that each sample is associated with a label distribution.   At present, the existing approaches are proposed for the single-view  LDL problem with labeled data,  while the multi-view LDL problem with labeled and unlabeled data has not been considered. In this paper, we  put forward the multi-view semi-supervised label distribution learning with local structure complementarity (MVSS-LDL) approach, which exploits the local nearest neighbor structure of each view and emphasizes the complementarity of    local nearest neighbor structures in multiple views.   Specifically speaking,  we first explore the local structure of  view $v$ by computing the $k$-nearest neighbors. As a result, the  $k$-nearest neighbor set  of each sample  $\boldsymbol{x}_i$  in view $v$ is attained. Nevertheless, this  $k$-nearest neighbor set describes only a part of the nearest neighbor information of sample  $\boldsymbol{x}_i$. In order to obtain a more comprehensive description of sample $\boldsymbol{x}_i$'s nearest neighbors, we complement the nearest neighbor set in view $v$ by incorporating sample  $\boldsymbol{x}_i$'s nearest neighbors in   other views.   Lastly, based on the complemented nearest neighbor set in each view,   a graph learning-based multi-view semi-supervised LDL model is constructed.    By considering the complementarity of local nearest neighbor structures,  different views can mutually provide the local structural information to complement each other. To the best of our knowledge, this is the first attempt at multi-view  LDL. Numerical studies have demonstrated that MVSS-LDL   attains explicitly better classification performance than the existing single-view LDL methods.

\end{abstract}


\section{Introduction}

\IEEEPARstart{L}{abel} Distribution Learning (LDL) \cite{new1,new2,new3} is an innovative learning framework that each sample is associated with a label distribution. The label distribution is a multi-dimensional vector, where each element is called the label description degree, representing the relevance to the corresponding label \cite{LDL}.  LDL aims to construct a learning model to map the samples to label distributions.  For example, in Fig. 1, the image is annotated with a label distribution, where  the label description degrees are 0.665 for “Mountain” (Mou),  0.213 for “Sky”, and 0.122 for “Cloud” (Clo); the other labels are 0.  It is seen that “Mountain” has a  larger label description degree than “Sky” and “Cloud”, since the image is mostly dominated by “Mountain”. LDL is different from multi-label learning. In multi-label learning,  sample $\boldsymbol{x}$ is annotated with a binary value (i.e., 0 or 1) to each possible label $y$,  representing whether  sample $\boldsymbol{x}$ is relevant to label $y$. In LDL,  sample $\boldsymbol{x}$ is annotated with a real value in [0, 1] to each possible label $y$, representing the relevance to label  $y$. LDL provides a more precise  description of the relevance to labels.
  To date, LDL has been widely applied in various tasks, e.g., age estimation, crowd counting  and emotion recognition \cite{newnew1,newnew2}.

\par At present, a considerable number of works have been done on LDL. For example, \cite{LDLF} extends the differentiable decision trees to solve LDL problems.  \cite{deepLDL} adopts a deep learning framework to leverage the label ambiguity in LDL data. \cite{GLLE} employs label enhancement   to  reconstruct the label distributions in the LDL tasks.  \cite{LDLLC} enhances the classification performance of LDL by leveraging the correlations between different labels.   \cite{ALDL} introduces an active  learning strategy for LDL, which identifies the informative samples based on the   disagreement of the committee members. \cite{RWLMLDL} presents the re-weighting large margin approach to address the discrepancy between the training process and testing process. It employs the $L_1$-loss and re-weighting scheme  to handle the inconsistency. 
 
\begin{figure}[t]
  \centering 
\includegraphics[width=9cm]{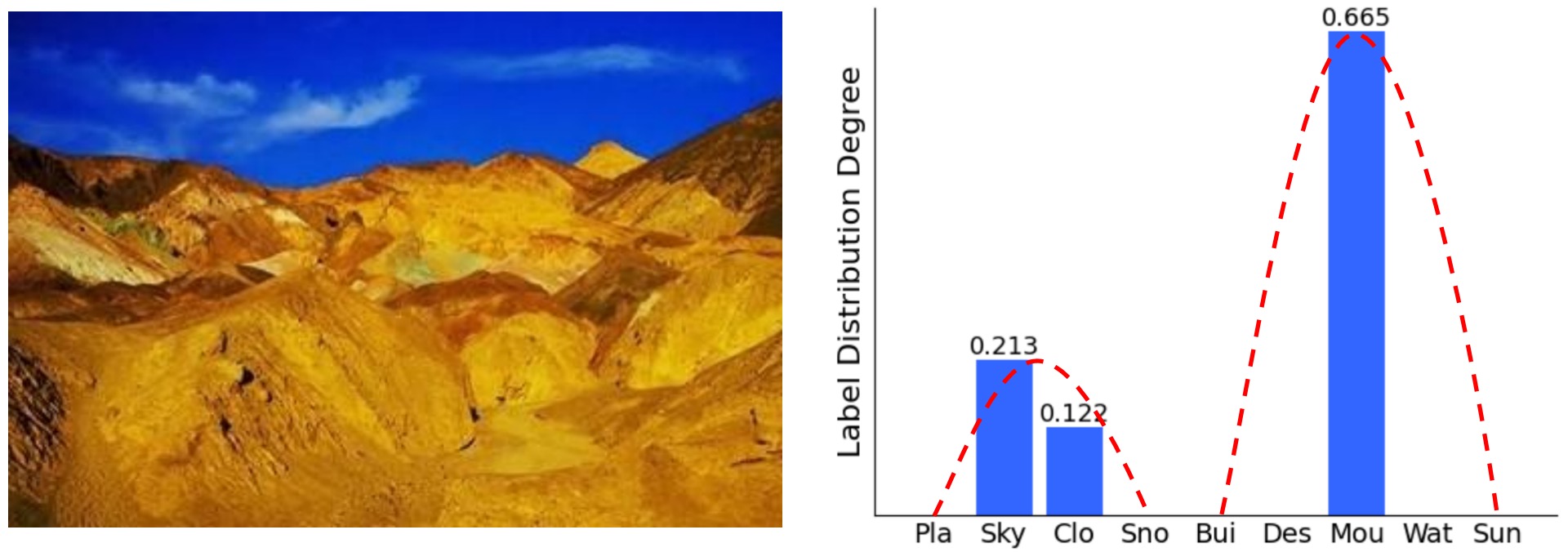}\\ 
\caption{An example of label distributions. The distribution degrees are 0.665 for “Mountain”  (Mou), 0.213 for “Sky”, and 0.122 for “Cloud”  (Clo). The other labels are 0. }  \vspace{-0.7cm}
\label{flowchart} 
\end{figure} 

\par Despite much progress on LDL,  the existing approaches are proposed for the single-view  LDL problem with labeled data,  while the multi-view LDL problem with labeled and unlabeled data has not been considered. On the one hand, in practice, a sample can be described by multiple views based on different feature extracting methods or data sources. For example, in image recognition,  the  image can be characterized by the shape  or texture features.  Moreover, in video analysis, a segment of video can be represented by the visual  or audio features.  From these examples, it is observed that each view describes a part of the sample, and we can incorporate these views to gain a more comprehensive understanding of the sample. Nevertheless, the existing LDL approaches focus on the single-view data, and the multi-view LDL problems have not been taken into account. On the other hand, apart from the labeled data, we may acquire some unlabeled data. Although this data is unannotated, it can be utilized to refine the LDL model learned on the labeled data. Therefore, how to incorporate the multi-view information and unlabeled data into building up a more accurate LDL model remains a key challenge.

\par  In this paper, we propose the multi-view semi-supervised label distribution learning with local structure complementarity (MVSS-LDL) approach, which explores the local nearest neighbor structure of each view and 
complements the local nearest neighbor structure of one view by integrating the nearest neighbor information of other views. Different from traditional multi-view learning, MVSS-LDL emphasizes the complementarity of local nearest neighbor structures in multiple views.  Firstly, we exploits the local structure of  view $v$ by computing the $k$-nearest neighbors. Thus,  the  $k$-nearest neighbor set of   sample  $\boldsymbol{x}_i$  in view $v$ is obtained. Secondly,
we complement the  nearest neighbor set in view $v$  by integrating the nearest neighbor information in other views. It is known that in multi-view learning, each sample $\boldsymbol{x}_i$ is represented by a number of views, and each view describes only a part of  sample $\boldsymbol{x}_i$. Likewise,  for sample $\boldsymbol{x}_i$, the $k$-nearest neighbor set  in view $v$  depicts only a part of  its nearest neighbor information. In order to obtain a more comprehensive description of sample $\boldsymbol{x}_i$'s nearest neighbors, we complement the nearest neighbor set in view $v$ by integrating sample  $\boldsymbol{x}_i$'s nearest neighbors in   other  views.   Lastly, we construct a   graph learning-based multi-view semi-supervised LDL model based on the complemented nearest neighbor set in each view.  By considering the complementarity of   nearest neighbor structures,  different views can mutually provide the local structural information to complement each other.   The experiments have validated that MVSS-LDL outperforms the existing single-view LDL approaches.

The contributions of  MVSS-LDL  are outlined below.

\begin{enumerate}
  \item[1)] We address the LDL problem with multi-view semi-supervised data. To the best of our knowledge, this is the first attempt at multi-view  LDL.
  \item[2)] Different from traditional multi-view learning,  we explore  the local nearest neighbor structure of each view and emphasize the complementarity of  local nearest neighbor structures  in multiple views.
  \item[3)]  Numerical studies have demonstrated that MVSS-LDL   attains explicitly better classification performance than the existing single-view LDL methods.
\end{enumerate}

\section{Related Work}

A considerable number of LDL approaches have been put forward. Based on the availability of data labels, these approaches are   categorized into two groups: supervised LDL and semi-supervised LDL. 

\par The supervised LDL approaches learn the  model  using only the labeled data. For example, \cite{LDL}  
solves the LDL problem using different strategies, e.g., converting  LDL  into traditional learning problem, and adapting the existing algorithms to resolve label distributions. \cite{deepLDL} transforms the single-label  problem and multi-label problem into LDL by assigning each sample a label distribution, and builds up a deep convolutional neural networks to learn the LDL model. 
\cite{LDLM} extends the margin theory to LDL  and designs an support vector regression-based LDL approach by incorporating  the adaptive margin loss.
\cite{LDLLC} explores the label relations by projecting the label relations onto a distance metric. The distance of labels is then measured based on this metric. 
\cite{LDL-LDM} investigates the label relations by conducting clustering on the training samples and  exploring the label distribution manifold of distinct clusters.  
\cite{LDL-LRR} introduces the ranking loss  to preserve the relationship of label ordering and combines it with Kullback-Leibler (KL) divergence.  \cite{ILDL} proposes the representation distribution alignment  approach to minimize the discrepancy between feature information and label information, such that the mismatch of the training data distribution and  testing data distribution can be diminished. 
\cite{FALDL} designs  the unimodal-concentrated loss for   adaptive LDL in  ordinal regression tasks. In \cite{FALDL}, three principles of adaptive LDL are proposed, i.e.,   highest probability at the ground-truth, unimodal distribution, and adaptive distribution for different samples. 
\par The semi-supervised LDL approaches construct the  model  using both the labeled data and unlabeled data. For instance, 
\cite{IncomLDL} deals with  the LDL problem with incomplete supervision information, and employs the trace norm minimization technique to investigate the relationship of label distributions.  \cite{RankMatch} adopts the averaging strategy, which  
 forms a pseudo label distribution  using the averaged predicting outputs of augmented data.
\cite{SSLDL-LCME}  utilizes the label  relationship   and   manifold information to estimate the label distribution of  unannotated data.
\cite{PGE-SLDL} constructs a graph  based on both the annotated data and unannotated data, and uses it to recover the label distribution matrix for unannotated data. It seeks an orthogonal mapping of the dataset to describe the local geometry. 
\cite{SALDL} combines  semi-supervised learning and  adaptation learning to solve the LDL problem with a small amount of annotated data. \cite{SLDL-CO} incorporates the $L_{2,1}$-norm and manifold regularization to exploit the implicit information in samples.

\par Although LDL has made great strides, the existing LDL  algorithms are proposed for single-view LDL that each sample is characterized by a single view. In practical tasks, a sample may be depicted by multiple  views. For instance, in image recognition, the image can be characterized from the shape or texture views. Combining these views can help us to obtain a more comprehensive description of the sample. However, the existing LDL methods do not take the multi-view data into account.  To handle the multi-view LDL data, MVSS-LDL is put forward, which can aggregate the nearest neighbor information in multiple views  to  boost the LDL model.

  \section{PROPOSED APPROACH}

\subsection{Preliminary}
Let $\mathcal{Y} = \{y_1, \cdots, y_q\}$ be the label space. Suppose that the LDL dataset is comprised of $l$ labeled samples, i.e., {($\boldsymbol{x}_1$, $\widetilde{\boldsymbol{d}}_1$), ($\boldsymbol{x}_2$, $\widetilde{\boldsymbol{d}}_2$), \dots, ($\boldsymbol{x}_l$, $\widetilde{\boldsymbol{d}}_l$)}, and $u$ unlabeled samples, i.e., {$\boldsymbol{x}_{l+1}$,  $\boldsymbol{x}_{l+2}$, \dots,  $\boldsymbol{x}_{n}$}. Here, $n=l+u$ is the total number of samples.  $\widetilde{\boldsymbol{d}}_i  \in \mathbb{R}^{q \times 1}$  is the label distribution of the labeled sample $\boldsymbol{x}_i$. It has $\widetilde{\boldsymbol{d}}_i$ = $[d_{i}^{1}, d_{i}^{2}, \dots, d_{i}^{q}]^\top$, where $d_{i}^{j}$ is the label description degree of sample $\boldsymbol{x}_i$ associated with label $y_j$. For these label description degrees, it has  ${d}_{i}^{j} \in [0, 1]$ and $\sum_{j=1}^q d_{i}^{j}=1$. Moreover,  each sample $\boldsymbol{x}_i$  is depicted from $V$ views and then represented by $V$ feature vectors, i.e., $\boldsymbol{x}_i^1, \boldsymbol{x}_i^2, \dots,    \boldsymbol{x}_i^V$, where $\boldsymbol{x}_i^v \in \mathbb{R}^{\rho_v}$ is the representation of $\boldsymbol{x}_i$ in  view $v$, and $\rho_v$ is the dimension number.   The basic notations   are shown in Table \ref{notations}.

\subsection{Graph-Based Multi-View Semi-Supervised LDL}
We introduce graph learning into constructing the multi-view semi-supervised LDL model. In graph learning, the similarity graph is a key step.  In order to learn the similarity graph, we define the similarity matrix  $\mathbf{S} \in \mathbb{R}^{nV \times nV}$, where the element $s_{ij}^v$ is the similarity of samples $\boldsymbol{x}_i^v$ and $\boldsymbol{x}_j^v$.  For the similarity matrix  $\mathbf{S}$, the learning problem is  
\begin{equation}\label{init S}
\begin{aligned}
&\min_{\mathbf{S}}\sum_{v=1}^{V}\sum_{i=1}^{n}\big\|\boldsymbol{x}_{i}^{v}-\sum_{j\in \mathcal{N}(\boldsymbol{x}_{i}^{v})}s_{ij}^{v}\boldsymbol{x}_{j}^{v}\big\|_{2}^{2} \\
&s.t. \ \mathbf{S} \mathbf{J}_{nV  \times V} =\mathbf{1}_{nV \times V}, \ \ \mathbf{S}\geq\mathbf{0}, \\
& \quad \ \; s_{ij}^{v}\geq0 \ (\forall j \in \mathcal{N}(\boldsymbol{x}_{i}^{v})), \   \ s_{ij}^{v}=0  \ (\forall j \notin \mathcal{N}(\boldsymbol{x}_{i}^{v})), \\
\end{aligned}
\end{equation}
 
\noindent where $\mathbf{J}_{nV  \times V}$   is a matrix that for the $i$-th column, the  $((i-1)V+1)$-th to $(iV)$-th elements are 1, and the other elements are 0;  $\mathbf{1}_{nV \times V}$ is an one matrix with $nV$ rows and $V$ columns. The constraint  $\mathbf{S} \mathbf{J}_{nV  \times V} =\mathbf{1}_{nV \times V}$ implies that for a sample $\boldsymbol{x}_{i}$, the sum of similarity values  in each view should be equal to 1, i.e., $\sum_{j=1}^{n} s_{ij}^{v}=1$.  Moreover,   $\mathcal{N}(\boldsymbol{x}_i^v)$ is a subset that contains  the sub-scripts of sample $\boldsymbol{x}_{i}$'s $k$-nearest neighbors ($k$NN) in  view $v$. The last set of constraints  indicates that if $\boldsymbol{x}^v_j$  (i.e., $ j \in \mathcal{N}(\boldsymbol{x}_{i}^{v})$)  is among the $k$-nearest neighbors of  $\boldsymbol{x}_i^v$, it has $s^v_{ij} \geq 0$. Otherwise, $s^v_{ij} = 0$ holds. It is seen that the $k$-nearest neighbor $\boldsymbol{x}_j^v$ (i.e., $ j \in \mathcal{N}(\boldsymbol{x}_{i}^{v})$) is weighted by the similarity value $s_{ij}^{v}$, and problem (\ref{init S}) focuses on minimizing the difference of each sample and its similarity-weighted nearest neighbors.

Define $\mathbf{D} \in \mathbb{R}^{nV \times q}$ as the predicted label  matrix, which is comprised of the predicted label distributions $\boldsymbol{d}^v_i$. Here, $\boldsymbol{d}^v_i$ is the predicted label distribution of sample $\boldsymbol{x}^v_i$.
Intuitively, if sample $\boldsymbol{x}^v_j$ lies close to $\boldsymbol{x}^v_i$, they may have similar label distributions, i.e.,  $\boldsymbol{d}^v_j$ being close to $\boldsymbol{d}^v_i$. Thus, for the  matrix $\mathbf{D}$, the learning problem is

\begin{equation}\label{init D}
\begin{aligned}
&\min_{\mathbf{D}} \sum_{v=1}^{V}\sum_{i=1}^{n}\big\|\boldsymbol{d}_{i}^{v}-\sum_{j\in \mathcal{N}(\boldsymbol{x}_{i}^{v})}s_{ij}^{v}\boldsymbol{d}_{j}^{v}\big\|_{2}^{2} \\
&s.t. \ \mathbf{D}\mathbf{1}_{q}=\mathbf{1}_{nV}, \ \  \mathbf{D}\geq \mathbf{0}, \\
& \quad \ \; \mathbf{I}_{lV\times nV}\mathbf{D}=\tilde{\mathbf{D}}_{l}, \\
\end{aligned}
\end{equation}

\noindent where $\mathbf{1}_{q}$ and $\mathbf{1}_{nV}$ are one vectors. The constraint $\mathbf{D}\mathbf{1}_q = \mathbf{1}_{nV}$ indicates that for each label distribution $\boldsymbol{d}_i^v$, the sum of its elements should be equal to 1. Furthermore,  $\mathbf{I}_{lV \times nV}$ is a matrix with the diagonal elements being 1 and the other elements being 0. $\widetilde{\mathbf{D}}_l \in \mathbb{R}^{lV \times q}$ is the  ground-truth label  matrix that contains the ground-truth label distributions of the labeled samples  $\boldsymbol{x}^v_i$ ($i=1, \dots, l$). Hence, the constraint $\mathbf{I}_{lV \times nV} \mathbf{D} = \widetilde{\mathbf{D}}_l$ makes the predicted label  matrix $\mathbf{D}$  equal to the ground-truth label matrix $\widetilde{\mathbf{D}}_l$ for the labeled samples.

\begin{table}[t]
\caption{SUMMARY OF BASIC NOTATIONS}
\centering
\label{notations}
\renewcommand\arraystretch{1.25}
\begin{tabular}{ll}
  \toprule
  Notations & Mathematical Meanings \\ \hline
   $\boldsymbol{x}_i^v$ & The representation of sample $\boldsymbol{x}_i$ in view $v$  \\
   $ \widetilde{\boldsymbol{x}}_i^v $ & Redefined sample for  $\boldsymbol{x}_i^v$ \\
    $q$ & Number of labels  \\
   $l, u$ & Number of labeled and unlabeled samples    \\
  $ \widetilde{\boldsymbol{d}_i}$ & Ground-truth label distribution of  labeled sample $\boldsymbol{x}_i^v$ \\
  $\boldsymbol{d}_i^v$ & Predicted label distribution of  sample $\boldsymbol{x}_i^v$ \\
  $s_{ij}^v$ & Similarity weight of samples $\boldsymbol{x}_i^v$ and $\boldsymbol{x}_j^v$ \\
  $\tilde{\mathbf{D}}_{l}$   & Ground-truth  label distribution matrix, containing  $ \widetilde{\boldsymbol{d}_i}$ \\
   $ \mathbf{D}$ & Predicted label distribution matrix, containing $\boldsymbol{d}_i^v$ \\
   $\mathbf{S}$ & Similarity weight matrix, containing $s_{ij}^v$  \\
  $\mathbf{W}^{v}$   & Model parameter of  view $v$ \\
   $\mathbf{W}$  & Concatenated matrix of $\mathbf{W}^{v}$ \\
 $\mathcal{N}(\boldsymbol{x}_i^v)$ & Subset  containing  the sub-scripts of   $\boldsymbol{x}_i$'s $k$NN
  in  view $v$ \\
$\mathcal{N}(\boldsymbol{x}_i)$ & Subset  containing  the sub-scripts of   $\boldsymbol{x}_i$'s $k$NN in  all views \\
$ n_i $ & The number of sub-scripts in the subset $ \mathcal{N}(\boldsymbol{x}_i)$\\
$ n $ &  The total number of training samples ($n=l+u$) \\
  \bottomrule
\end{tabular}  
\end{table} 

In multi-view learning, each sample $\boldsymbol{x}_i$ can be represented by multiple feature vectors, i.e., $\boldsymbol{x}_i^1, \boldsymbol{x}_i^2, \dots, \boldsymbol{x}_i^V$. These feature vectors are associated with the same sample $\boldsymbol{x}_i$ and thus should have the same label distribution.  Thus, we employ the label distribution consistency regularizer, as follows. 
\begin{equation}\label{add label consistency}
\sum_{v,u=1}^V  \sum_{i=1}^{n}\left\|\boldsymbol{d}_{i}^{v}-\boldsymbol{d}_{i}^{u}\right\|_{2}^{2},  
\end{equation}
\noindent where $\boldsymbol{d}^v_i$ and $\boldsymbol{d}^u_i$ are the label distributions of   $\boldsymbol{x}_i$ in view $v$ and view $u$, respectively. Since $\boldsymbol{d}^v_i$ and $\boldsymbol{d}^v_i$ are related to the same sample $\boldsymbol{x}_i$ and should have the same
value, we minimize the difference of $\boldsymbol{d}^v_i$ and $\boldsymbol{d}^u_i$ to make them close to each other.

Moreover,  according to the consistency principle \cite{multiview1,multiview2}, different views may share a common underlying representation, which can be learned by exploiting the consistency between views. For this sake, we present the similarity consistency regularizer, which is be defined as:
\begin{equation}\label{add feature consistency}
\sum_{v, u=1}^V \sum_{i=1}^{n}\sum_{j\in \mathcal{N}(\boldsymbol{x}_i^v)}(s_{ij}^v-s_{ij}^u)^2,
\end{equation}

\noindent where $s^v_{ij}$ is the similarity of $\boldsymbol{x}_i$ and $\boldsymbol{x}_j$ in view $v$, and $s^u_{ij}$ is the similarity of $\boldsymbol{x}_i$ and $\boldsymbol{x}_j$ in view $u$. Since different views may share a common underlying representation, we minimize the difference of $s^v_{ij}$ and $s^u_{ij}$, such that sample $\boldsymbol{x}_i$ can have similar relationship to its $k$-nearest neighbors in different views.

By incorporating  (\ref{init S}) - (\ref{add feature consistency}), the
learning problem   is given by: \vspace{-0.2cm}
\begin{equation}\label{OF}
\begin{aligned}
&\min_{\mathbf{S},\mathbf{D},\mathbf{W}} \lambda \sum_{v=1}^{V}\sum_{i=1}^{n}\big\|(\mathbf{W}^{v})^{\top}\boldsymbol{x}_{i}^{v}- \boldsymbol{d}_{i}^v \big\|_{2}^{2} +\sum_{v=1}^{V}\left\|\mathbf{W}^{v}\right\|_{F}^{2} \ \ \ \  \ \\
\end{aligned}
\end{equation} \vspace{-0.1cm}
\begin{equation} 
\begin{aligned}
&+\mu_1\sum_{v=1}^V\sum_{i=1}^{n}\big \|\boldsymbol{x}_i^v-\sum_{j\in \mathcal{N}(\boldsymbol{x}_i^v)}s_{ij}^v\boldsymbol{x}_j^v\big\|_2^2 \nonumber \\
&+\mu_2\sum_{v=1}^V\sum_{i=1}^{n}\big \|\boldsymbol{d}_i^v-\sum_{j\in \mathcal{N}(\boldsymbol{x}_i^v)}s_{ij}^v\boldsymbol{d}_j^v\big\|_2^2  \nonumber \\
&+\sigma\sum_{v,u=1}^{V} \sum_{i=1}^{n}\sum_{j\in \mathcal{N}(\boldsymbol{x}_{i}^v)}(s_{ij}^{v}-s_{ij}^{u})^{2} +\gamma\sum_{v,u=1}^V \sum_{i=1}^{n}\left\|\boldsymbol{d}_i^v-\boldsymbol{d}_i^u\right\|_2^2 \nonumber \\
&s.t. \ \ \mathbf{S} \mathbf{J}_{nV  \times V} =\mathbf{1}_{nV \times V}, \ \ \mathbf{S}\geq \mathbf{0}, \nonumber \\
&\quad \ \ \  \mathbf{D}\mathbf{1}_q=\mathbf{1}_{nV}, \ \  \mathbf{I}_{lV\times nV}\mathbf{D}=\mathbf{\widetilde{D}}_l, \ \ \mathbf{D}\geq \mathbf{0},
\end{aligned}
\end{equation}

\noindent where $\lambda$, $\mu_1$,   $\mu_2$,  $\sigma$, and  $\gamma$ are nonnegative parameters; $\mathbf{W}^v \in \mathbb{R}^{\rho_v \times q}$ is the model parameter of  view $v$.

\subsection{Local Nearest Neighbor Structure Complementarity}

 In multi-view learning, each  sample may be depicted from distinct views and transformed into a number of feature vectors. Thus, multi-view learning has two essential principles, i.e., the consistency principle and complementarity principle \cite{multiview1,multiview2}. The consistency principle states that the representations of the same sample ought to be assigned to the same class. 
The complementarity principle states that each view describes a part of the sample, and a more comprehensive description of the sample is attained by combining multiple views. However, problem (\ref{OF}) considers only the consistency principle, and the complementarity principle has not been taken into account. In problem (\ref{OF}),  the label distribution consistency regularizer and similarity consistency regularizer  consider  the consensus information between views, and only the consistency principle is realized. 

In order to realize both the consistency principle and complementarity principle, we complement the local nearest neighbor structure  of one view by incorporating the nearest neighbor information of other views, such that different views can mutually provide the local structural information to enrich each other. The learning problem of   MVSS-LDL   is given by: \vspace{-0.3cm}

\begin{equation}\label{OF_all}
\begin{aligned}
&\min_{\mathbf{S},\mathbf{D},\mathbf{W}}  \lambda \sum_{v=1}^{V}\sum_{i=1}^{n}\big\|(\mathbf{W}^{v})^{\top}\boldsymbol{x}_{i}^{v}- \boldsymbol{d}_{i}^v \big\|_{2}^{2} +\sum_{v=1}^{V}\left\|\mathbf{W}^{v}\right\|_{F}^{2} \ \ \ \  \ \\
\end{aligned}
\end{equation} \vspace{-0.1cm}
\begin{equation} 
\begin{aligned}
&+\mu_1\sum_{v=1}^V\sum_{i=1}^{n}\big \|\boldsymbol{x}_i^v-\sum_{j\in \mathcal{N}(\boldsymbol{x}_i )}s_{ij}^v\boldsymbol{x}_j^v\big\|_2^2 \nonumber \\
&+\mu_2\sum_{v=1}^V\sum_{i=1}^{n} \big \|\boldsymbol{d}_i^v - \sum_{j\in \mathcal{N}(\boldsymbol{x}_i )}s_{ij}^v\boldsymbol{d}_j^v\big\|_2^2  \nonumber \\
&+\sigma\sum_{v,u=1}^{V} \sum_{i=1}^{n}\sum_{j\in \mathcal{N}(\boldsymbol{x}_{i} )}(s_{ij}^{v}-s_{ij}^{u})^{2} +\gamma\sum_{v,u=1}^V \sum_{i=1}^{n}\left\|\boldsymbol{d}_i^v-\boldsymbol{d}_i^u\right\|_2^2 \nonumber \\
&s.t. \ \ \mathbf{S} \mathbf{J}_{nV  \times V} =\mathbf{1}_{nV \times V}, \ \ \mathbf{S}\geq \mathbf{0}, \nonumber \\
&\quad \ \ \ \mathbf{D}\mathbf{1}_q=\mathbf{1}_{nV}, \ \ \mathbf{I}_{lV\times nV}\mathbf{D}=\mathbf{\widetilde{D}}_l, \ \ \mathbf{D}\geq \mathbf{0},
\end{aligned}
\end{equation}

\noindent where it has $\mathcal{N}(\boldsymbol{x}_i)= \mathcal{N}(\boldsymbol{x}_i^1) \cup \mathcal{N}(\boldsymbol{x}_i^2) \cup \dots \cup \mathcal{N}(\boldsymbol{x}_i^V)$. It is a subset that contains the sub-scripts of sample $\boldsymbol{x}_i$’s $k$-nearest neighbors in all views.

The differences of problems (\ref{OF}) and   (\ref{OF_all}) exist in the third, fourth and fifth terms. In these terms, $ \mathcal{N}(\boldsymbol{x}_i^v)$ is changed into  $\mathcal{N}(\boldsymbol{x}_i)$. Here, $ \mathcal{N}(\boldsymbol{x}_i^v)$ is the nearest neighbor set in view $v$, which  contains  sample $\boldsymbol{x}_i$'s $k$-nearest neighbors only in view $v$. Distinctively, $\mathcal{N}(\boldsymbol{x}_i)$ is the complemented nearest neighbor set, which involves sample $\boldsymbol{x}_i$'s $k$-nearest neighbors in all views.  It can be seen that we complement the nearest neighbor set in view $v$ by incorporating sample  $\boldsymbol{x}_i$'s nearest neighbors in other views, such that   a more comprehensive description of sample $\boldsymbol{x}_i$'s nearest neighbors can be obtained.

Let us take the third term as an example.  In problem (\ref{OF}), the third term is $\ ||\boldsymbol{x}_i^v - \sum_{ j \in \mathcal{N}(x_i^v)} s_{ij}^v \boldsymbol{x}_j^v||^2_2$. Here, $ \mathcal{N}(\boldsymbol{x}_i^v)$  contains  sample $\boldsymbol{x}_i$'s $k$-nearest neighbors only in view $v$. Thus, this term is employed to minimize the difference of sample $\boldsymbol{x}_i$ and its $k$-nearest neighbors in view $v$. 
Nevertheless, it considers only the $k$-nearest neighbors in view $v$, while the nearest neighbor information of other views cannot be utilized to complement the classifier  of view $v$. To handle this problem, we modify it to be  $ ||\boldsymbol{x}_i^v - \sum_{ j \in \mathcal{N}(\boldsymbol{x}_i)} s_{ij}^v \boldsymbol{x}_j^v||^2_2$, as shown in the third term of problem (\ref{OF_all}). Here,  $\mathcal{N}(\boldsymbol{x}_i)= \mathcal{N}(\boldsymbol{x}_i^1) \cup \mathcal{N}(\boldsymbol{x}_i^2)$ $\cup$  $\dots$ $\cup$  $\mathcal{N}(\boldsymbol{x}_i^V)$  contains sample $\boldsymbol{x}_i$'s $k$-nearest neighbors   in all views. By employing this term,  the  nearest neighbor information of other views is treated  as the complementary information of view $v$, and  complements the classifier  in view $v$.   Similar cases happen to the fourth and fifth terms.

\subsection{Optimization}

Problem (\ref{OF_all}) contains three sets of unknown variables, i.e., $s^v_{ij}$, $\boldsymbol{d}^v_{i}$ and $\mathbf{W}^v$. To solve this problem,  the alternative optimization technique is employed. Specifically, we optimize one set of the variables and fix the remaining variables in each turn. This process iterates until the algorithm converges.\vspace{+0.2cm}

$Update$ $S$ by fixing $\mathbf{D}$ and $\mathbf{W}$. When $\mathbf{D}$ and $\mathbf{W}$ are fixed, the first, second and last terms in   (\ref{OF_all}) can be removed, and the learning problem contains only the unknown variables $s_{ij}^v$. Considering that the similarity value $s_{ij}^v$ is only related to $\boldsymbol{x}_{i}^v$ and its $k$-nearest neighbors, we can optimize the similarity values of each sample independently, and problem (\ref{OF_all}) can be split into $n$ sub-problems. The sub-problem for optimizing the similarity values of sample $\boldsymbol{x}_i$ is as follows. 

\begin{equation}\label{UpdateS_i}
\begin{aligned}
&\min_{\widetilde{\mathbf{S}}_{i}} \ \mu_1 \sum_{v=1}^{V}\big\|\boldsymbol{x}_{i}^{v}-\sum_{j\in \mathcal{N}(\boldsymbol{x}_{i})}s_{ij}^{v}\boldsymbol{x}_{j}^{v}\big\|_{2}^{2} \\
& \quad +\mu_2 \sum_{v=1}^{V}\big\|\boldsymbol{d}_{i}^{v}-\sum_{j\in \mathcal{N}(\boldsymbol{x}_{i}^{v})}s_{ij}^{v}\boldsymbol{d}_{j}^{v}\big\|_{2}^{2} \\
& \quad +\sigma\sum_{v,u=1}^{V} \sum_{j\in \mathcal{N}(\boldsymbol{x}_{i})}(s_{ij}^{v}-s_{ij}^{u})^{2} \\
& s.t. \   \widetilde{\mathbf{S}}_i \mathbf{J}_{n_iV  \times V} =\mathbf{1}_{V}^\top, \ \ \widetilde{\mathbf{S}}_i\geq\mathbf{0},
\end{aligned}
\end{equation}

\noindent where $\widetilde{\mathbf{S}}_i \in \mathbb{R}^{1  \times  Vn_i}$ contains the similarity value $s_{ij}^v$ of sample $\boldsymbol{x}_i$ in all views, and $ n_i $ is the number of sub-scripts in the subset $ \mathcal{N}(\boldsymbol{x}_i)$.  $ \mathbf{J}_{n_iV  \times V}$ is a matrix that for the $i$-th column, the  $((i-1)V+1)$-th to $(iV)$-th elements are 1, and the other elements are 0.   $\mathbf{1}_V^\top$ is an $V$-dimensional row vector with all elements being 1.

For sample $\boldsymbol{x}_i$, we re-index the nearest neighbor  $\boldsymbol{x}_j^v$, label distribution  $\boldsymbol{d}_j^v$ and similarity value  $s_{ij}^v$ as $\boldsymbol{x}_{\mathcal{N}_i(j)}^v$, $\boldsymbol{d}_{\mathcal{N}_i(j)}^v$ and $s_{\mathcal{N}_i(j)}^v$, respectively. Here, $\boldsymbol{x}_{\mathcal{N}_i(j)}^v$ is the $j$-th nearest neighbor of sample $\boldsymbol{x}_i^v$, and $\boldsymbol{d}_{\mathcal{N}_i(j)}^v$ is label distribution of $\boldsymbol{x}_{\mathcal{N}_i(j)}^v$.  $s_{\mathcal{N}_i(j)}^v$ is the similarity value between $\boldsymbol{x}_i^v$ and the $j$-th nearest neighbor  $\boldsymbol{x}_{\mathcal{N}_i(j)}^v$. Based on these, the similarity vector is rewritten as $\widetilde{\mathbf{S}}_i$ $=$ $[ s_{\mathcal{N}_i(1)}^1$,  $\dots$, $s_{\mathcal{N}_i(n_i)}^1$, $\dots$, $s_{\mathcal{N}_i(1)}^V$,   $\dots$, $s_{\mathcal{N}_i(n_i)}^V]$ $\in$ $ \mathbb{R}^{1  \times  Vn_i}$. Thus, problem (\ref{UpdateS_i}) is changed into:

\begin{equation}\label{update_S_all}
\begin{aligned}
\min_{\widetilde{\mathbf{S}}_i} \ & \mu_1\sum_{v=1}^V||\boldsymbol{x}_i^v-\sum_{j=1}^{n_i}s_{\mathcal{N}_{i(j)}}^v \boldsymbol{x}_{\mathcal{N}_{i(j)}}^v||_2^2 \\
& + \mu_2\sum_{v=1}^V||\boldsymbol{d}_i^v-\sum_{j=1}^{n_i}s_{\mathcal{N}_{i(j)}}^v \boldsymbol{d}_{\mathcal{N}_{i(j)}}^v||_2^2 \\
& +  \sigma  \sum_{v,u=1}^V \sum_{j=1}^{n_i} (s_{\mathcal{N}_{i(j)}}^u - s_{\mathcal{N}_{i(j)}}^v)^2 \\
 s.t.  & \ \widetilde{\mathbf{S}}_i \mathbf{J}_{n_iV  \times V} =\mathbf{1}_{V}^\top, \ \ \widetilde{\mathbf{S}}_i\geq\mathbf{0}.
\end{aligned}
\end{equation}

\par We redefine the  vector $\widetilde{\boldsymbol{x}}_i^v$ $=$ $[\mathbf{0}_{\rho_1}$, $ \dots$, $\mathbf{0}_{\rho_{v-1}}$, $\boldsymbol{x}_i^v$, $\mathbf{0}_{\rho_{v+1}}$, $\dots$, $\mathbf{0}_{\rho_V}] \in \mathbb{R}^{\rho \times 1}$, where the $v$-th element is $\boldsymbol{x}_i^v$, and the other elements are zeros vectors;  $\rho=\rho_1+\rho_2+\cdots+\rho_V$ is the total number of dimensions in all views. Moreover, denote  $\mathbf{D}^{\boldsymbol{x}_i} = [\mathbf{D}^{\boldsymbol{x}_i}_1, \mathbf{D}^{\boldsymbol{x}_i}_2, \dots, \mathbf{D}^{\boldsymbol{x}_i}_V]^\top \in \mathbb{R}^{Vn_i \times \rho}$, where it has $\mathbf{D}^{\boldsymbol{x}_i}_v=[\widetilde{\boldsymbol{x}}_i^v-\widetilde{\boldsymbol{x}}^v_{\mathcal{N}_i(1)},  \widetilde{\boldsymbol{x}}_i^v- \widetilde{\boldsymbol{x}}^v_{\mathcal{N}_i(2)}, \dots,
\widetilde{\boldsymbol{x}}^v_i-\widetilde{\boldsymbol{x}}^v_{\mathcal{N}_i(n_i)}]^\top  \in \mathbb{R}^{n_i \times \rho}$. Let $ \mathbf{D}^{i}$ $=$ $[\mathbf{D}^{i}_1, \mathbf{D}^{i}_2, \dots, \mathbf{D}^{i}_V ]^\top  \in \mathbb{R}^{Vn_i \times q}$, where it has $ \mathbf{D}^{i}_v=[\boldsymbol{d}_i^v-\boldsymbol{d}_{\mathcal{N}_i(1)}^v, \boldsymbol{d}_i^v-\boldsymbol{d}_{\mathcal{N}_i(2)}^v, \dots, 
\boldsymbol{d}_i^v-\boldsymbol{d}_{\mathcal{N}_i(n_i)}^v]^\top  \in \mathbb{R}^{n_i \times q}$.  By redefining    $\mathbf{D}^{\boldsymbol{x}_i}$ and $\mathbf{D}^{i}$, the first and second terms in   (\ref{update_S_all})  are changed into $ \widetilde{\mathbf{S}}_i \mathbf{G}^{\boldsymbol{x}_i} (\widetilde{\mathbf{S}}_i)^\top $ and $ \widetilde{\mathbf{S}}_i \mathbf{G}^{i} (\widetilde{\mathbf{S}}_i)^\top $, respectively. Here, it has $\mathbf{G}^{\boldsymbol{x}_i}=\mathbf{D}^{\boldsymbol{x}_i}(\mathbf{D}^{\boldsymbol{x}_i})^\top$ and $\mathbf{G}^{i}=\mathbf{D}^{i}(\mathbf{D}^{i})^\top$. 

Define $\boldsymbol{e}_i=[\mathbf{0}_{n_i \times n_i}$, $\dots$, $\mathbf{0}_{n_i \times n_i}$, $\mathbf{I}_{n_i \times n_i}$, $\mathbf{0}_{n_i \times n_i}$, $\dots$, $\mathbf{0}_{n_i \times n_i}]^\top$ $\in$ $\mathbb{R}^{Vn_i \times n_i}$, which has $V$ elements with the $i$-th element being the identity matrix $\mathbf{I}_{n_i \times n_i}$ and the others being the zero matrices $\mathbf{0}_{n_i \times n_i}$. Based on  $\boldsymbol{e}_i$, we   have
\begin{equation}\label{}
\begin{aligned}
& \quad \ \  \sum_{v,u=1}^V \sum_{j=1}^{n_i} (s_{\mathcal{N}_{i(j)}}^u - s_{\mathcal{N}_{i(j)}}^v)^2 \\
&=\sigma\sum_{\stackrel{v,u=1}{v<u}}^{V} \widetilde{\mathbf{S}}_{i} (\mathbf{e}_{v}-\mathbf{e}_{u})(\mathbf{e}_{v}-\mathbf{e}_{u})^{\top} (\widetilde{\mathbf{S}}_{i})^{\top},
\end{aligned}
\end{equation}

\noindent where $v < u$ is to prevent redundant calculations. Thus, problem (\ref{update_S_all}) can be rewritten as \vspace{-0.2cm}
\begin{equation}
\begin{aligned}
& \min_{\widetilde{\mathbf{S}}_{i}} \widetilde{\mathbf{S}}_{i}  \Big[\mu_1 \mathbf{G}^{\boldsymbol{x}_{i}}+ \mu_2 \mathbf{G}^{i} +\sigma\sum_{v,u=1\atop v<u}^{V}(\mathbf{e}_{v}-\mathbf{e}_{u})(\mathbf{e}_{v}-\mathbf{e}_{u})^{\top} \Big]   (\widetilde{\mathbf{S}}_{i})^{\top} \nonumber
\end{aligned}
\end{equation}\vspace{-0.2cm}
\begin{equation} \label{UpdateS_OF}
\begin{aligned}
  s.t. \ \widetilde{\mathbf{S}}_i \mathbf{J}_{n_iV  \times V} =\mathbf{1}_{V}^\top, \ \widetilde{\mathbf{S}}_{i}\geq \mathbf{0}. \ \ \ \ \ \ \ \ \ \ \ \ \ \ \ \ \ \ \ \ \ \  
\end{aligned}
\end{equation}

The optimization problem (\ref{UpdateS_OF}) is a standard quadratic programming (QP) problem with $Vn_i$ variables, and we can resolve it via the existing QP tools. Once each $\widetilde{\mathbf{S}}_i$ is solved, we can update the similarity   matrix $\mathbf{S}$. \vspace{+0.2cm}

$Update$ $D$ by fixing $\mathbf{S}$ and $\mathbf{W}$. With fixed $\mathbf{S}$ and $\mathbf{W}$, the optimization problem (\ref{OF_all}) is changed into  \vspace{-0.1cm}
\begin{equation}\label{update_D}
\begin{aligned}
& \min_{ \mathbf{D} } \lambda \sum_{v=1}^{V}\sum_{i=1}^{n}\big\|(\mathbf{W}^{v})^{\top}\boldsymbol{x}_{i}^{v}-\boldsymbol{d}^v_{i}\big\|_{2}^{2} \ \ \ \ \ \ \ \ \ \ \ \ \ \ \ \ \ \ \ 
\end{aligned}
\end{equation} \vspace{-0.1cm}
\begin{equation}
\begin{aligned}
&  +\mu_2 \sum_{v=1}^{V}\sum_{i=1}^{n}\big\|\boldsymbol{d}_{i}^{v}-  \sum_{j\in \mathcal{N}(\boldsymbol{x}_i )}s_{ij}^v\boldsymbol{d}_j^v
 \big\|_{2}^{2} +\gamma\sum_{v,u=1}^{V} \sum_{i=1}^{n}\big\|\boldsymbol{d}_{i}^{v}-\boldsymbol{d}_{i}^{u}\big\|_{2}^{2} \\
& s.t. \ \mathbf{D}\mathbf{1}_{q}=\mathbf{1}_{nV}, \ \  \mathbf{D}\geq \mathbf{0}, \\ \nonumber
& \quad \ \ \mathbf{I}_{lV\times nV}\mathbf{D}=\mathbf{\widetilde{D}}_{l}. \nonumber
\end{aligned}
\end{equation}

Before optimizing   $\mathbf{D}$, we  redefine the   matrix $\mathbf{W}$ $=$ $[\mathbf{W}^1$, $\mathbf{W}^2$, $\dots$, $\mathbf{W}^V]  \in \mathbb{R}^{\rho \times q}$ and the vector $\widetilde{\boldsymbol{x}}_i^v$ $=$ $[\mathbf{0}_{\rho_1}$, $ \dots$, $\mathbf{0}_{\rho_{v-1}}$, $\boldsymbol{x}_i^v$, $\mathbf{0}_{\rho_{v+1}}$, $\dots$, $\mathbf{0}_{\rho_V}] \in \mathbb{R}^{\rho \times 1}$. Hence, it has $(\mathbf{W}^{v})^{\top}\boldsymbol{x}_{i}^{v}=\mathbf{W}^\top \widetilde{\boldsymbol{x}}_i^v$. In addition,   denote $\mathbf{H} =[\mathbf{H}^1, \mathbf{H}^2, \dots, \mathbf{H}^V]^\top \in \mathbb{R}^{nV \times q}$, where it has $\mathbf{H}^v=[\mathbf{W}^\top \widetilde{\boldsymbol{x}}_1^v, \dots, \mathbf{W}^\top \widetilde{\boldsymbol{x}}_{n}^v]^\top \in \mathbb{R}^{n  \times q}$. Let  $\widehat{\boldsymbol{d}} = vec(\mathbf{D}) \in \mathbb{R}^{nVq \times 1}$ and $\widehat{\boldsymbol{h}} = vec(\mathbf{H}) \in \mathbb{R}^{nVq \times 1}$, where $vec(\bullet)$ is the vectorization operator. Based on these, the first term of problem (\ref{update_D}) is changed into:
\begin{equation}\label{D-1}
\begin{aligned}
& \quad \lambda \sum_{v=1}^{V}\sum_{i=1}^{n}\big\|(\mathbf{W}^{v})^{\top}\boldsymbol{x}_{i}^{v}-\boldsymbol{d}^v_{i}\big\|_{2}^{2}\\
&= \lambda \Big( \widehat{\boldsymbol{d}}^\top\widehat{\boldsymbol{d}}-2\widehat{\boldsymbol{h}}^\top\widehat{\boldsymbol{d}}  \Big) + const,
\end{aligned}
\end{equation}

\noindent where $const=  \lambda \widehat{\boldsymbol{h}}^\top\widehat{\boldsymbol{h}}$ is a constant.

Denote the  matrix $\mathbf{L}_{D}=\mathbf{I}_{nV \times nV} - \mathbf{S}$, where $\mathbf{I}_{nV \times nV}$ is an identity matrix; $\mathbf{S} \in \mathbb{R}^{nV \times nV}$ is the similarity matrix learned in the above step, i.e., the step of optimizing $\mathbf{S}$.  Moreover, let   $\mathbf{\Lambda}_1   \in \mathbb{R}^{nVq \times nVq}$ be a diagonal matrix, in which the diagonal element is  $ \mathbf{L}_{D}^{\top}  \mathbf{L}_{D}$.  Thus, the  second term of problem (\ref{update_D}) can be changed into

\begin{equation}\label{D-2}
\begin{aligned}
& \quad \mu_2 \sum_{v=1}^V \sum_{i=1}^{n}\big\|\boldsymbol{d}_i^v -   \sum_{j\in \mathcal{N}(\boldsymbol{x}_i )}s_{ij}^v\boldsymbol{d}_j^v  \big\|_2^2 \\
&=\mu_2 \ tr(\mathbf{D}^{\top} \mathbf{L}_{D}^{\top}  \mathbf{L}_{D}  \mathbf{D}) \\
&=\mu_2 \ \widehat{\boldsymbol{d}}^{\top}\mathbf{\Lambda}_{1}\widehat{\boldsymbol{d}}.
\end{aligned}
\end{equation}

  The last term of problem (\ref{update_D}) can be reformulated as:

\begin{equation}\label{D-3}
\begin{aligned}
& \quad \gamma  \sum_{v,u=1}^{V}\sum_{i=1}^{n}\big\|\boldsymbol{d}_{i}^{v}-\boldsymbol{d}_{i}^{u}\big\|_{2}^{2}\\
& = \gamma tr(\mathbf{D}^{\top}\mathbf{L}_{Q}\mathbf{D})\\
& = \gamma \widehat{\boldsymbol{d}}^{\top}\boldsymbol{\Lambda}_{2} \widehat{\boldsymbol{d}},
\end{aligned}
\end{equation}

\noindent where $tr(\cdot)$ is the trace of a matrix, and $\mathbf{L}_{Q}$ is a graph Laplacian matrix. It has $\mathbf{L}_{Q}=\mathbf{G}_Q -  \frac{Q + Q^\top}{2} \in \mathbb{R}^{nV \times nV}$, and $\mathbf{G}_Q$ is a diagonal matrix with the   diagonal element being $\mathbf{G}_{Q_{ii}} = \sum_{j} \frac{Q_{ij}+{Q_{ji}}}{2}$. Here,  $Q_{ij}=1$ holds if the  label distributions $\boldsymbol{d}_{i}^{v}$ and $\boldsymbol{d}_{j}^{u}$ are associated with the same sample, i.e., $i=j$.  Otherwise,   $Q_{ij}=0$ holds.  $\mathbf{\Lambda}_2 \in \mathbb{R}^{nVq \times nVq}$ is a  diagonal matrix with the   diagonal element being 
$\mathbf{L}_{Q}$. 

By substituting (\ref{D-1}) - (\ref{D-3}),   problem (\ref{update_D}) is transformed into: \vspace{-0.2cm}
\begin{equation} 
\begin{aligned}
&\min_{\widehat{\boldsymbol{d}}} \widehat{\boldsymbol{d}}^\top
 \big( \lambda \mathbf{I}_{nVq\times nVq}+ \mu_2 \Lambda_1 
 + \gamma\Lambda_2\big) \widehat{\boldsymbol{d}}
 -2 \lambda \widehat{\boldsymbol{h}}^\top\widehat{\boldsymbol{d}} + const \nonumber
 \end{aligned}
\end{equation}\vspace{-0.3cm}
\begin{equation}\label{update_D_OF}
\begin{aligned}
&\ s.t. \ \mathbf{D}\mathbf{1}_{q}=\mathbf{1}_{nV}, \ \ \mathbf{D}\geq\mathbf{0}, \ \ \ \ \ \ \ \ \ \ \ \ \ \ \ \ \  \ \ \ \ \ \ \ \ \ \ \ \ \ \\
&\quad \ \ \ \ \mathbf{I}_{lV\times nV}\mathbf{D}=\widetilde{\mathbf{D}}_{l}.       \ \ \ \ \ \ \ \
\end{aligned}
\end{equation}

Problem (\ref{update_D_OF}) is a standard QP problem, and we can resolve it via the existing QP tools. \vspace{+0.2cm}

$Update \  W$ by fixing $\mathbf{S}$ and $\mathbf{D}$. When $\mathbf{S}$ and $\mathbf{D}$ are fixed, the learning problem of $\mathbf{W}$ is given by 
\begin{equation}\label{update_W}
\min_{ \mathbf{W}}  \lambda \sum_{v=1}^{V}\sum_{i=1}^{n}\big\|(\mathbf{W}^{v})^{\top}\boldsymbol{x}_{i}^{v}-\boldsymbol{d}^v_{i}\big\|_{2}^{2} + \sum_{v=1}^{V}\left\|\mathbf{W}^{v}\right\|_{F}^{2}.
\end{equation}

 By redefining  $\mathbf{W}$ $=$ $[\mathbf{W}^1$, $\mathbf{W}^2$, $\dots$, $\mathbf{W}^V]  \in \mathbb{R}^{\rho \times q}$ and $\widetilde{\boldsymbol{x}}_i^v$ $=$ $[\mathbf{0}_{\rho_1}$, $ \dots$, $\mathbf{0}_{\rho_{v-1}}$, $\boldsymbol{x}_i^v$, $\mathbf{0}_{\rho_{v+1}}$, $\dots$, $\mathbf{0}_{\rho_V}] \in \mathbb{R}^{\rho \times 1}$, it has $(\mathbf{W}^{v})^{\top}\boldsymbol{x}_{i}^{v}=\mathbf{W}^{\top} \widetilde{\boldsymbol{x}}_i^v$ and $\sum \left\|\mathbf{W}^{v}\right\|_{F}^{2} =\left\|\mathbf{W} \right\|_{F}^{2} $. 
 Based on these, the closed-form solution  of $\mathbf{W}$ is as follows.

\begin{equation}\label{Update_W}
\mathbf{W}=  ( \lambda^2 \mathbf{X}^\top\mathbf{X}+  \lambda \mathbf{I}_{\rho  \times \rho})^{-1}\mathbf{X}^\top\mathbf{D}.
\end{equation}

\noindent where  it has $\mathbf{X} = [\widetilde{\boldsymbol{x}}_1^1, \dots, \widetilde{\boldsymbol{x}}_n^1, \dots, \widetilde{\boldsymbol{x}}_1^V, \dots, \widetilde{\boldsymbol{x}}_n^V ]^\top   \in \mathbb{R}^{nV \times \rho} $, $\mathbf{D} = [d_1^1, \dots, d_n^1, \dots, d_1^V, \dots, d_n^V ]^\top   \in \mathbb{R}^{nV \times q} $, and $\mathbf{I}_{\rho \times \rho}$ is an identity matrix.

These steps are iterated alternatively until the convergence is achieved.  For a test sample $\boldsymbol{x}_t =\{\boldsymbol{x}^1_t, \boldsymbol{x}^2_t, \dots, \boldsymbol{x}^V_t \}$, its label distribution  $\boldsymbol{d}_t$ is predicted as follows. Firstly,  redefine $\boldsymbol{x}_t^v$ as $\widetilde{\boldsymbol{x}}^v_t$. Secondly, compute the label distribution in each view, i.e., $\boldsymbol{d}^v_t=\mathbf{W}^\top \widetilde{\boldsymbol{x}}^v_t$. Lastly, the final perdition of the test sample $\boldsymbol{x}_t$ is given by $\boldsymbol{d}_t= \frac{1}{V} \sum_{v=1}^{V} \boldsymbol{d}^v_t$. The pseudo-code of MVSS-LDL  is summarized in Algorithm \ref{pseudo}.

\section{Experiment}
To verify the effectiveness, MVSS-LDL is validated  on real-life LDL datasets.    The experimental environment is MacOS 10.15 with Intel Core i7-9750H CPU and 32GB RAM.

\subsection{Evaluation Metrics}
We employ six evaluation metrics, including Chebyshev, Clark, KL divergence, Canberra, Cosine, and Intersection.  Among these metrics, Chebyshev, Clark, KL divergence and Canberra measure the distance of label distributions, where lower values correspond to  superior results. Moreover, Cosine and Intersection measure the similarity of label distributions, where higher values correspond to  superior results. The definition of these metrics is illustrated in Table \ref{metric}, where “↓” implies that a lower value is desired, and  “↑” implies that a higher value is desired.   $\boldsymbol{p}$ and $\boldsymbol{q}$ represent the label distributions; $p_i$ and $q_i$ represent the $i$-th element.  

 \begin{figure*}[htbp]	
    \centering
    \subfloat[SCUT-FBP]{
        \includegraphics[width=0.18\linewidth]{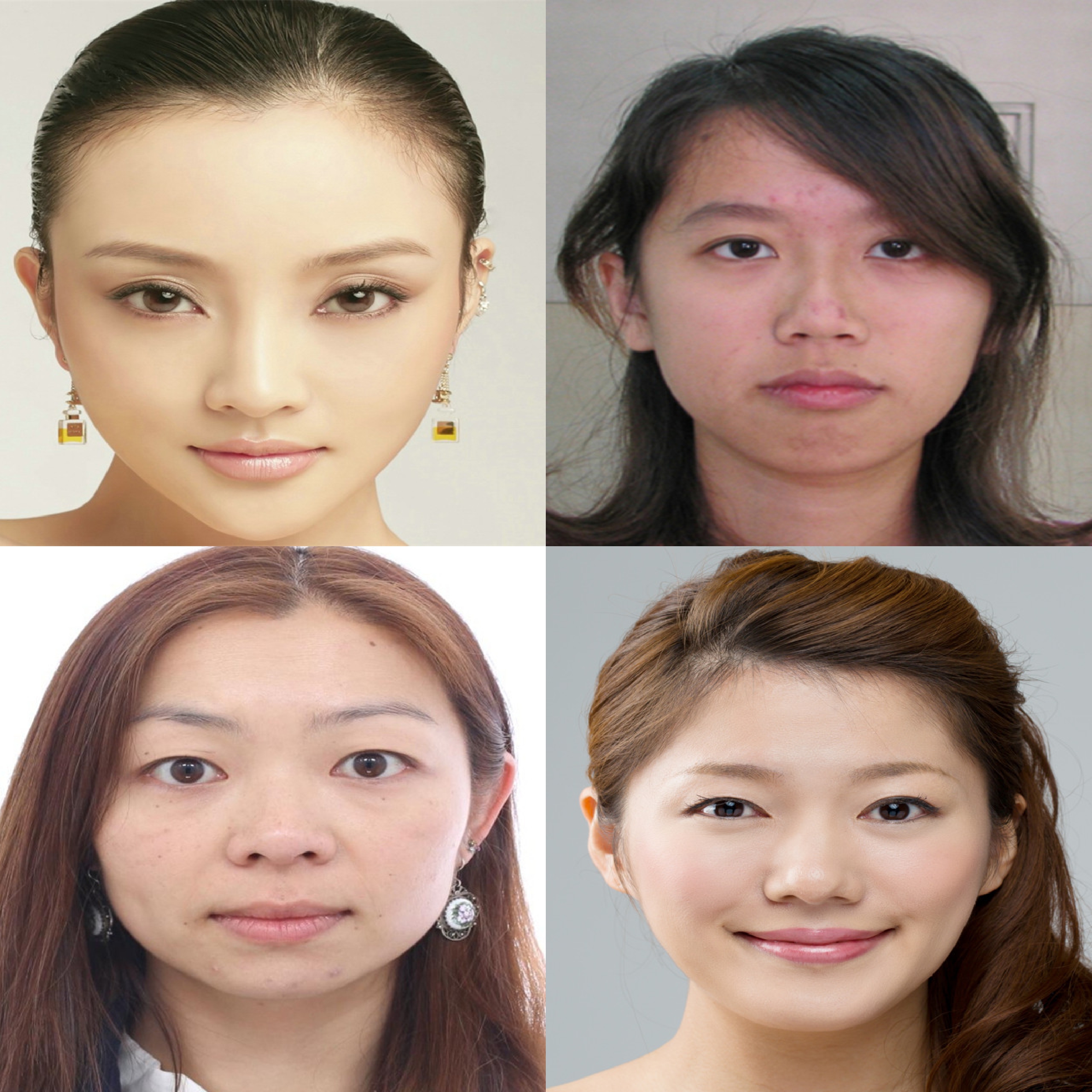}
        } 
    \subfloat[Emotion6]{
        \includegraphics[width=0.18\linewidth]{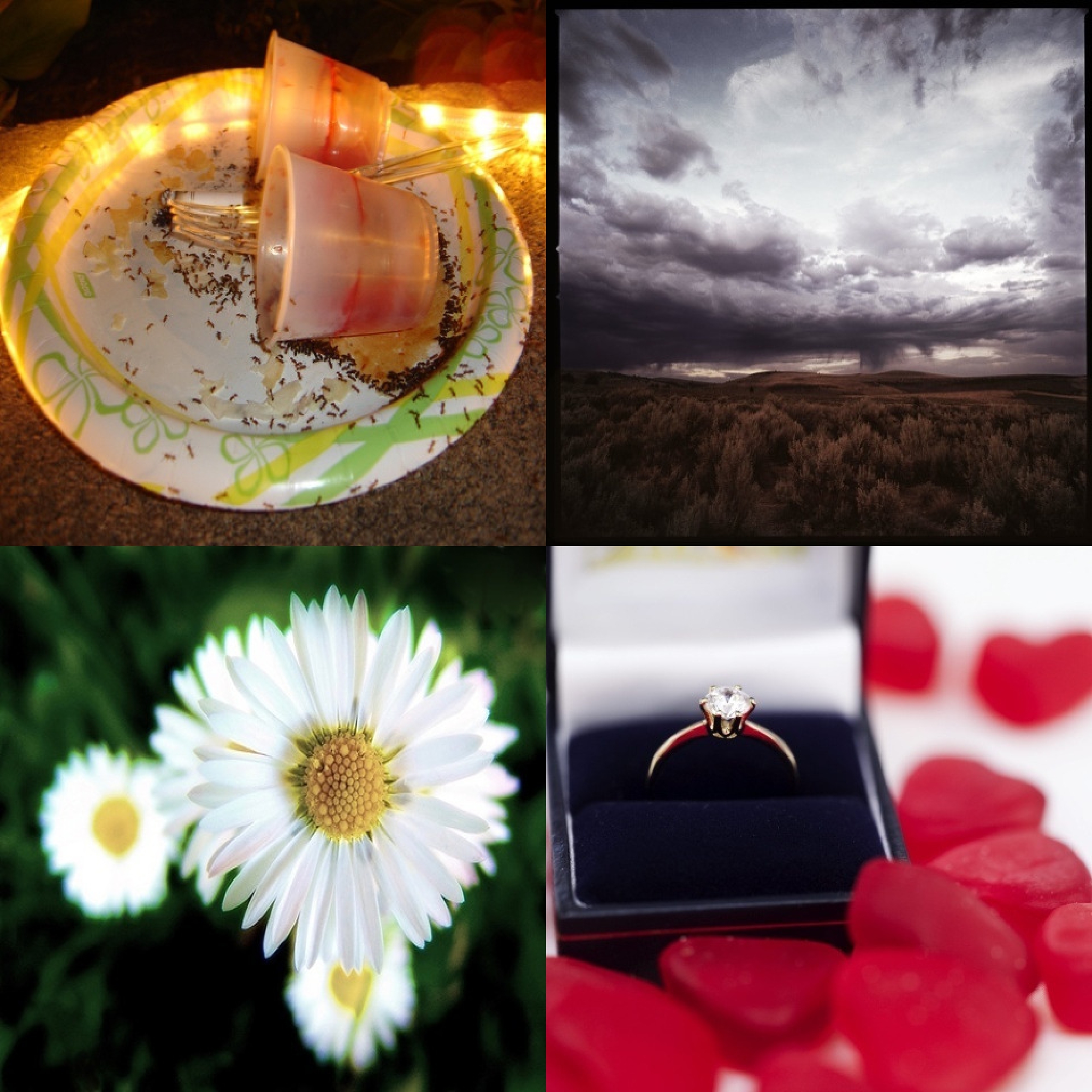}
        } 
        \subfloat[fbp5500]{
        \includegraphics[width=0.18\linewidth]{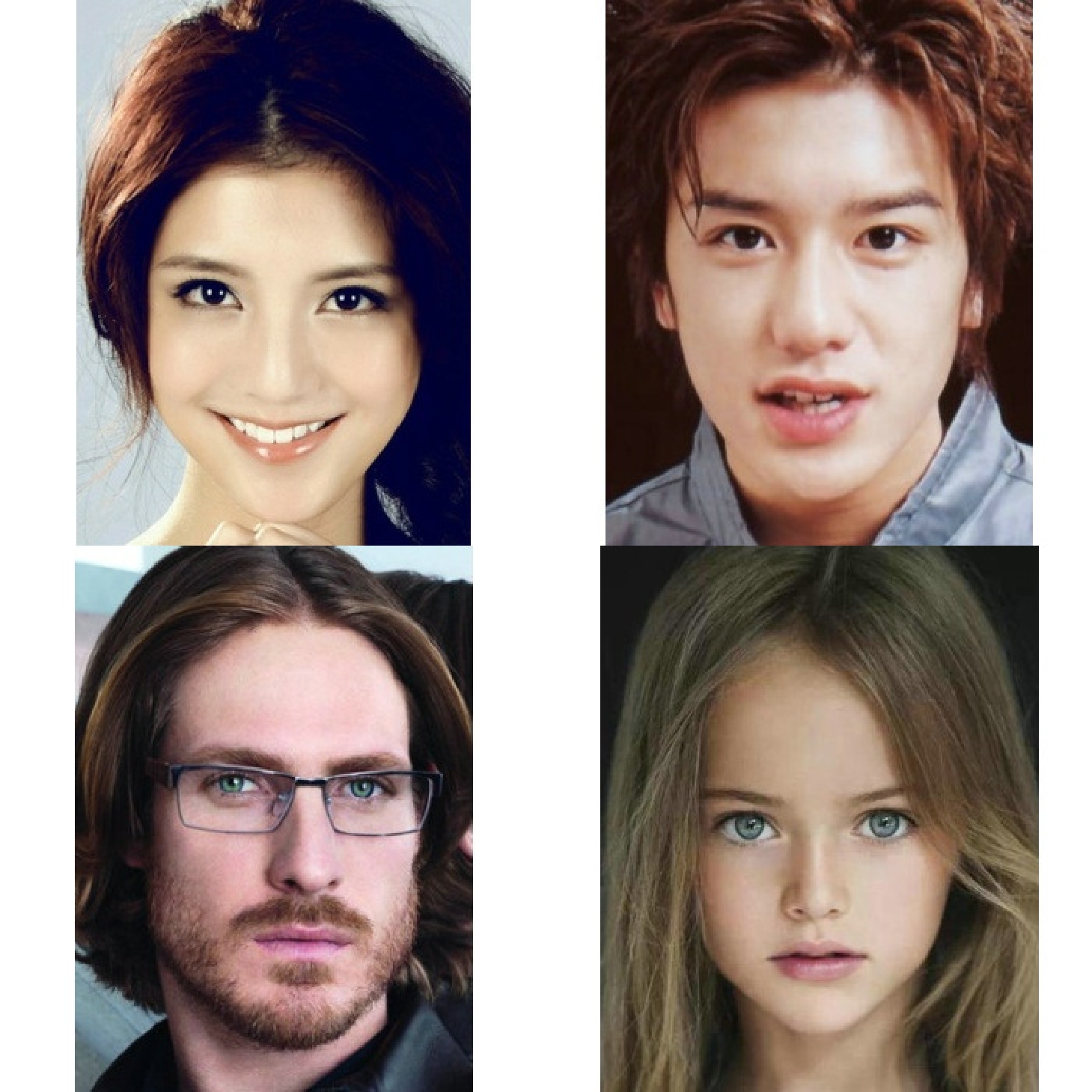}
        }
        \subfloat[Twitter-LDL]{
        \includegraphics[width=0.18\linewidth]{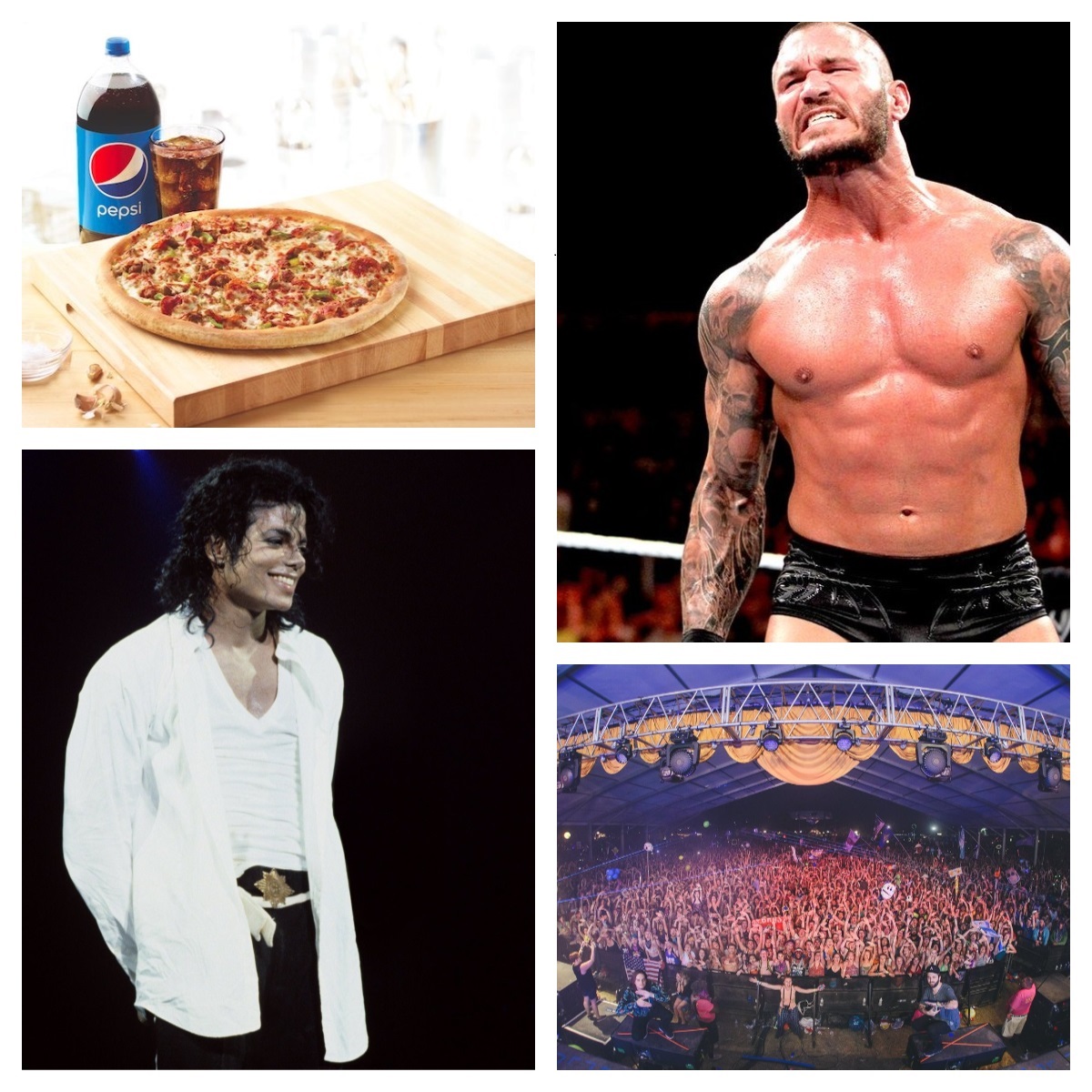}
        } 
        \subfloat[Flickr-LDL]{
        \includegraphics[width=0.18\linewidth]{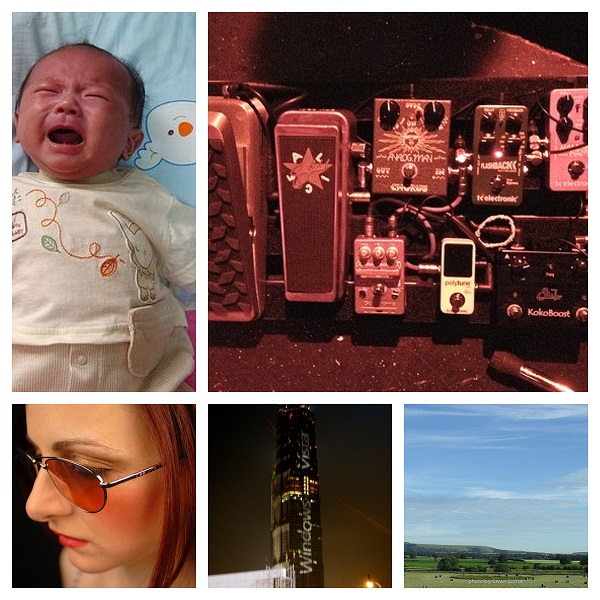}
        } 
      
        \caption{Sample images from (a) SCUT-FBP, (b) Emotion6, (c) fbp5500,  (d) Twitter-LDL, and (e) Flickr-LDL. }
        \label{Dataset}

\end{figure*}

\begin{algorithm} [t]
\caption{ MVSS-LDL}\label{alg:cap}
\label{pseudo}
\begin{algorithmic}
\STATE   {\textbf{Input:} Labeled data $\{(\boldsymbol{x}_i^1, \boldsymbol{x}_i^2,...,\boldsymbol{x}_i^V, \widetilde{\boldsymbol{d}}_i) \}_{i=1}^l$ and unlabeled data $\{ (\boldsymbol{x}_i^1, \boldsymbol{x}_i^2,...,\boldsymbol{x}_i^V) \}_{i=l+1}^{n}$; nearest neighbors number $k$, regularization parameters $\mu_1$, $\mu_2$,  $\lambda$, $\sigma$ and $\gamma$.}
\STATE {\textbf{Process:}}
\STATE  {\ 1. \ Compute the $k$-nearest neighbors of each sample;}
\STATE  {\ 2. \ Initialize the similarity matrix $\mathbf{S}$ via  (\ref{init S});}
\STATE {\ 3. \ Initialize the label distribution matrix $\mathbf{D}$ via  (\ref{init D});}
\STATE {\ 4. \ \textbf{While} not converged}
\STATE {\ 5.  \ \ \ \ Compute the matrix $\mathbf{W}$ via  (\ref{Update_W});}
\STATE {\ 6.  \ \ \ \ \textbf{for}  $i$ = 1 \textbf{to} $n$ \textbf{do}}
\STATE {\ 7.  \ \ \ \ \ \ \ Obtain the vector $\widetilde{\mathbf{S}_i}$  via  (\ref{UpdateS_OF}) and update $\widetilde{\mathbf{S}_i}$  in the}
\STATE {\ \ \ \ \ \ \ \ \  \ \ similarity matrix $\mathbf{S}$;}
\STATE {\ 8. \ \ \ \  \textbf{end for}}
\STATE {\ 9. \ \ \ \ Calculate the label distribution matrix $\mathbf{D}$ via  (\ref{update_D_OF});}
\STATE {\ 10.  \textbf{End While}}
\STATE {\ 11. For the test sample $\boldsymbol{x}_t$, its label distribution  is predicted }
\STATE {\ \ \ \ \ as  $\boldsymbol{d}_t= \frac{1}{V} \sum_{v=1}^{V}  \mathbf{W}^\top \widetilde{\boldsymbol{x}}^v_t$;}
\STATE {\textbf{Output:} The predicted label distribution $\boldsymbol{d}_t$.} 
\end{algorithmic}  
\end{algorithm}

\begin{table}[htbp]
\renewcommand\arraystretch{1.6}
\caption{EVALUATION METRICS FOR LDL} 
\label{metric}
\begin{tabular}{ll} 
\toprule 
 LDL Metric ($abbr.$) & Formula \\
\midrule 
Chebyshev distance $(chebyshev)$ $\downarrow$ & $\max_{i}|p_{i}-q_{i}|$ \\
Clark distance $(clark)$ $\downarrow$ & $\sqrt{\sum_i\left(p_i-q_i\right)^2/\left(p_i+q_i\right)^2}$ \\
Canberra distance $(canberra)$ $\downarrow$ & $\sum_i|p_i-q_i|/p_i+q_i$ \\ 
KL divergence $(KL)$ $\downarrow$ & $\sum_ip_i\ln\frac{p_i}{q_i}$ \\
Cosine similarity $(cosine)$ $\uparrow$ & $\frac{\boldsymbol{p}^\top \boldsymbol{q}}{||\boldsymbol{p}||_2||\boldsymbol{q}||_2}$ \\
Intersection similarity $(intersection)$ $\uparrow$ & $\sum_i \min (p_i,q_i)$ \\
\bottomrule 
\end{tabular}
\end{table}

\subsection{Dataset Description}

The numerical validation is conducted on six multi-view LDL datasets, i.e., SCUT-FBP, Emotion6, fbp5500, RAF-ML, Twitter-LDL  and Flickr-LDL. These datasets are widely used in LDL works \cite{GLLE} - \cite{LDL-LRR}. Sample images are shown in Fig \ref{Dataset}. \vspace{+0.2cm}

\begin{table*}[]
\setlength{\tabcolsep}{10pt}
\renewcommand\arraystretch{1.3}
\setlength{\tabcolsep}{3.0mm}
\caption{Experimental results   with 10\% labeled  samples.}
\label{result}
\begin{tabular}{cccccccc}
\hline
Dataset                          & Method                         & Chebyshev$\downarrow$              & Clark$\downarrow$                 & Canberra$\downarrow$                & KL Divergence$\downarrow$           & Cosine$\uparrow$                  & Intersection $\uparrow$           \\ \hline
\multicolumn{1}{c|}{}            & \multicolumn{1}{c|}{MVSS-LDL}  & \textbf{0.3180$\pm$0.0023} & \textbf{1.3857$\pm$0.0019} & \textbf{2.6806$\pm$0.0024} & 0.6317$\pm$0.0041 & \textbf{0.7367$\pm$0.0025} & \textbf{0.5668$\pm$0.0024} \\
\multicolumn{1}{c|}{}            & \multicolumn{1}{c|}{LDL-LRR}   & 0.4640$\pm$0.0060          & 1.6830$\pm$0.0037          & 3.3778$\pm$0.0089          & 1.3989$\pm$0.0082          & 0.5164$\pm$0.0067          & 0.4612$\pm$0.0083          \\
\multicolumn{1}{c|}{}            & \multicolumn{1}{c|}{LDL-SCL}   & 0.7267$\pm$0.0048          & 1.9080$\pm$0.0042          & 3.9309$\pm$0.0086          & 8.3496$\pm$0.0056          & 0.3559$\pm$0.0052          & 0.2546$\pm$0.0089          \\
\multicolumn{1}{c|}{}            & \multicolumn{1}{c|}{LDL-LDM}   & 0.7394$\pm$0.0046          & 2.0956$\pm$0.0052          & 4.6888$\pm$0.0076          & 14.4140$\pm$0.0086         & 0.3357$\pm$0.0065          & 0.2302$\pm$0.0058          \\
\multicolumn{1}{c|}{SCUT-FBP}    & \multicolumn{1}{c|}{LDM-Incom} & 0.4343$\pm$0.0055          & 1.5309$\pm$0.0074          & 3.0920$\pm$0.0062          & 1.1838$\pm$0.0036          & 0.5173$\pm$0.0060          & 0.4483$\pm$0.0054          \\
\multicolumn{1}{c|}{}            & \multicolumn{1}{c|}{S2LDL}     & 0.3892$\pm$0.0045          & 1.4359$\pm$0.0045          & 2.7961$\pm$0.0094          & 0.7512$\pm$0.0067          & 0.7090$\pm$0.0061          & 0.4921$\pm$0.0057          \\
\multicolumn{1}{c|}{}            & \multicolumn{1}{c|}{IncomLDL}  & 0.3456$\pm$0.0043          & 1.4212$\pm$0.0061          & 2.7615$\pm$0.0074          & \textbf{0.6099$\pm$0.0039}         & 0.6990$\pm$0.0058          & 0.5298$\pm$0.0086          \\
 \hline
\multicolumn{1}{c|}{}            & \multicolumn{1}{c|}{MVSS-LDL}  & \textbf{0.3438$\pm$0.0024} & \textbf{1.6521$\pm$0.0013} & \textbf{3.7421$\pm$0.0020} & \textbf{0.6899$\pm$0.0033} & \textbf{0.6704$\pm$0.0028} & \textbf{0.5401$\pm$0.0039} \\
\multicolumn{1}{c|}{}            & \multicolumn{1}{c|}{LDL-LRR}   & 0.4493$\pm$0.0059          & 1.9549$\pm$0.0014          & 4.7213$\pm$0.0047          & 1.5770$\pm$0.0030          & 0.5257$\pm$0.0033          & 0.4433$\pm$0.0095          \\
\multicolumn{1}{c|}{}            & \multicolumn{1}{c|}{LDL-SCL}   & 0.4779$\pm$0.0021          & 1.9593$\pm$0.0049          & 4.4403$\pm$0.0048          & 1.7975$\pm$0.0043          & 0.6123$\pm$0.0062          & 0.4812$\pm$0.0041          \\
\multicolumn{1}{c|}{}            & \multicolumn{1}{c|}{LDL-LDM}   & 0.7592$\pm$0.0045          & 2.0520$\pm$0.0051          & 4.1300$\pm$0.0053          & 13.1710$\pm$0.0037         & 0.4149$\pm$0.0048          & 0.2999$\pm$0.0037          \\
\multicolumn{1}{c|}{Emotion6}    & \multicolumn{1}{c|}{LDM-Incom} & 0.3675$\pm$0.0036          & 1.7611$\pm$0.0046          & 4.0845$\pm$0.0055          & 0.9328$\pm$0.0042          & 0.5858$\pm$0.0057          & 0.4943$\pm$0.0052          \\
\multicolumn{1}{c|}{}            & \multicolumn{1}{c|}{S2LDL}     & 0.4923$\pm$0.0058          & 1.7220$\pm$0.0045          & 3.9634$\pm$0.0051          & 2.9634$\pm$0.0069          & 0.5255$\pm$0.0039          & 0.4252$\pm$0.0038          \\
\multicolumn{1}{c|}{}            & \multicolumn{1}{c|}{IncomLDL}  & 0.3887$\pm$0.0043          & 1.9032$\pm$0.0053          & 4.0832$\pm$0.0056          & 1.1324$\pm$0.0050          & 0.6390$\pm$0.0053          & 0.5261$\pm$0.0070          \\
 \hline
\multicolumn{1}{c|}{}            & \multicolumn{1}{c|}{MVSS-LDL}  & \textbf{0.3647$\pm$0.0013} & \textbf{1.4824$\pm$0.0031} & \textbf{2.8897$\pm$0.0044} & \textbf{0.6676$\pm$0.0042} & \textbf{0.7055$\pm$0.0024} & \textbf{0.5394$\pm$0.0028} \\
\multicolumn{1}{c|}{}            & \multicolumn{1}{c|}{LDL-LRR}   & 0.4128$\pm$0.0058          & 1.6227$\pm$0.0064          & 2.9653$\pm$0.0053          & 0.7904$\pm$0.0054          & 0.6701$\pm$0.0062          & 0.5024$\pm$0.0066         \\
\multicolumn{1}{c|}{}            & \multicolumn{1}{c|}{LDL-SCL}   & 0.4658$\pm$0.0043          & 1.6625$\pm$0.0037          & 3.3408$\pm$0.0094          & 1.7345$\pm$0.0038          & 0.6169$\pm$0.0088          & 0.5037$\pm$0.0071          \\
\multicolumn{1}{c|}{}            & \multicolumn{1}{c|}{LDL-LDM}   & 0.6866$\pm$0.0087          & 2.0524$\pm$0.0072          & 4.1296$\pm$0.0067          & 13.1714$\pm$0.0067         & 0.4149$\pm$0.0083          & 0.2999$\pm$0.0065          \\
\multicolumn{1}{c|}{fbp5500}     & \multicolumn{1}{c|}{LDM-Incom} & 0.3996$\pm$0.0074          & 1.5456$\pm$0.0073          & 3.0460$\pm$0.0088          & 0.8763$\pm$0.0058          & 0.6157$\pm$0.0066         & 0.4902$\pm$0.0079          \\
\multicolumn{1}{c|}{}            & \multicolumn{1}{c|}{S2LDL}     & 0.5723$\pm$0.0028          & 1.5220$\pm$0.00424          & 3.0029$\pm$0.0031          & 1.0656$\pm$0.0045          & 0.5322$\pm$0.00255          & 0.4522$\pm$0.0091          \\
\multicolumn{1}{c|}{}            & \multicolumn{1}{c|}{IncomLDL}  & 0.3649$\pm$0.0084          & 1.5114$\pm$0.0060          & 2.9513$\pm$0.0074          & 0.7342$\pm$0.0051          & 0.6756$\pm$0.0075          & 0.4929$\pm$0.0099          \\
 \hline
\multicolumn{1}{c|}{}            & \multicolumn{1}{c|}{MVSS-LDL}  & \textbf{0.3688$\pm$0.0037}         & \textbf{1.6000$\pm$0.0029}          & \textbf{3.4785$\pm$0.0033}          & \textbf{0.6788$\pm$0.0027}          & \textbf{0.6848$\pm$0.0031}          & \textbf{0.5177$\pm$0.0051}          \\
\multicolumn{1}{c|}{}            & \multicolumn{1}{c|}{LDL-LRR}   & 0.5803$\pm$0.0069          & 2.0191$\pm$0.0093          & 4.5475$\pm$0.0055          & 4.5475$\pm$0.0098          & 0.5216$\pm$0.0040          & 0.3887$\pm$0.0074          \\
\multicolumn{1}{c|}{}            & \multicolumn{1}{c|}{LDL-SCL}   & 0.4128$\pm$0.0057          & 1.7542$\pm$0.0055          & 3.6741$\pm$0.0050          & 2.6310$\pm$0.0086          & 0.5843$\pm$0.0046          & 0.4689$\pm$0.0042          \\
\multicolumn{1}{c|}{}            & \multicolumn{1}{c|}{LDL-LDM}   & 0.5078$\pm$0.0068          & 2.0089$\pm$0.0068          & 4.5979$\pm$0.0089          & 2.2847$\pm$0.0048          & 0.5889$\pm$0.0059          & 0.4544$\pm$0.0089          \\
\multicolumn{1}{c|}{RAF-ML}      & \multicolumn{1}{c|}{LDM-Incom} & 0.3916$\pm$0.0061          & 1.6841$\pm$0.0069          & 3.7013$\pm$0.0070          & 0.9147$\pm$0.0064          & 0.6031$\pm$0.0075          & 0.4708$\pm$0.0054          \\
\multicolumn{1}{c|}{}            & \multicolumn{1}{c|}{S2LDL}     & 0.5405$\pm$0.0058          & 1.7336$\pm$0.0069          & 3.7765$\pm$0.0059          & 2.5621$\pm$0.0083          & 0.5364$\pm$0.0084          & 0.4231$\pm$0.0068          \\
\multicolumn{1}{c|}{}            & \multicolumn{1}{c|}{IncomLDL}  & 0.4249$\pm$0.0042          & 1.7312$\pm$0.0033          & 3.7323$\pm$0.0045          & 1.0260$\pm$0.0053          & 0.5755$\pm$0.0078          & 0.4638$\pm$0.0041          \\
 \hline
\multicolumn{1}{c|}{}            & \multicolumn{1}{c|}{MVSS-LDL}  & \textbf{0.6398$\pm$0.0016} & 2.3318$\pm$0.0034          & \textbf{6.2535$\pm$0.0031} & 2.1070$\pm$0.0025          & \textbf{0.5922$\pm$0.0030} & \textbf{0.3265$\pm$0.0046} \\
\multicolumn{1}{c|}{}            & \multicolumn{1}{c|}{LDL-LRR}   & 0.7193$\pm$0.0036          & 2.6434$\pm$0.0045          & 7.2966$\pm$0.0081          & 9.2461$\pm$0.0053          & 0.3204$\pm$0.0067          & 0.2705$\pm$0.0051          \\
\multicolumn{1}{c|}{}            & \multicolumn{1}{c|}{LDL-SCL}   & 0.7280$\pm$0.0039          & 2.5556$\pm$0.0082          & 6.9638$\pm$0.0031          & 7.8806$\pm$0.0044          & 0.3939$\pm$0.0034          & 0.2479$\pm$0.0085          \\
\multicolumn{1}{c|}{}            & \multicolumn{1}{c|}{LDL-LDM}   & 0.7114$\pm$0.0071          & 2.6684$\pm$0.0094          & 7.3509$\pm$0.0065          & 12.1420$\pm$0.0089         & 0.3210$\pm$0.0027          & 0.2758$\pm$0.0058          \\
\multicolumn{1}{c|}{Twitter-LDL } & \multicolumn{1}{c|}{LDM-Incom} & 0.7220$\pm$0.0026          & 2.3537$\pm$0.0051          & 6.5364$\pm$0.0094          & \textbf{1.5775$\pm$0.0022} & 0.4221$\pm$0.0045          & 0.2411$\pm$0.0064          \\
\multicolumn{1}{c|}{}            & \multicolumn{1}{c|}{S2LDL}     & 0.8297$\pm$0.0056          & 2.3524$\pm$0.0033          & 6.5140$\pm$0.0064          & 1.8991$\pm$0.0051          & 0.4096$\pm$0.0086          & 0.2107$\pm$0.0046          \\
\multicolumn{1}{c|}{}            & \multicolumn{1}{c|}{IncomLDL}  & 0.6814$\pm$0.0076          & \textbf{2.3126$\pm$0.0056} & 6.3867$\pm$0.0080          & 1.6540$\pm$0.0060          & 0.4575$\pm$0.0064          & 0.2805$\pm$0.0042          \\
 \hline
\multicolumn{1}{c|}{}            & \multicolumn{1}{c|}{MVSS-LDL}  & \textbf{0.6733$\pm$0.0020} & \textbf{2.3967$\pm$0.0038} & \textbf{6.5994$\pm$0.0027} & 2.6680$\pm$0.0028          & \textbf{0.5542$\pm$0.0032} & \textbf{0.2942$\pm$0.0040} \\
\multicolumn{1}{c|}{}            & \multicolumn{1}{c|}{LDL-LRR}   & 0.7461$\pm$0.0041          & 2.7522$\pm$0.0038          & 7.6580$\pm$0.0047          & 11.1154$\pm$0.0052         & 0.4098$\pm$0.0075          & 0.2511$\pm$0.0050          \\
\multicolumn{1}{c|}{}            & \multicolumn{1}{c|}{LDL-SCL}   & 0.7552$\pm$0.0048          & 2.5850$\pm$0.0033          & 7.0717$\pm$0.0074          & 8.3829$\pm$0.0022          & 0.3804$\pm$0.0039          & 0.2235$\pm$0.0040          \\
\multicolumn{1}{c|}{}            & \multicolumn{1}{c|}{LDL-LDM}   & 0.8081$\pm$0.0018          & 2.7114$\pm$0.0084          & 7.5289$\pm$0.0065          & 18.5787$\pm$0.0063         & 0.2102$\pm$0.0046          & 0.1802$\pm$0.0092          \\
\multicolumn{1}{c|}{Flickr-LDL}  & \multicolumn{1}{c|}{LDM-Incom} & 0.7193$\pm$0.0040          & 2.4302$\pm$0.0036          & 6.7147$\pm$0.0058          & \textbf{1.6161$\pm$0.0040}         & 0.4198$\pm$0.0031          & 0.2324$\pm$0.0082          \\
\multicolumn{1}{c|}{}            & \multicolumn{1}{c|}{S2LDL}     & 0.8908$\pm$0.0097          & 2.4472$\pm$0.0065          & 6.7602$\pm$0.0095          & 2.0070$\pm$0.0011          & 0.3978$\pm$0.0073          & 0.2165$\pm$0.0072          \\
\multicolumn{1}{c|}{}            & \multicolumn{1}{c|}{IncomLDL}  & 0.7082$\pm$0.0074          & 2.4221$\pm$0.0057          & 6.6821$\pm$0.0035          & 1.7081$\pm$0.0041 & 0.4248$\pm$0.0061          & 0.2465$\pm$0.0041          \\
 \hline
\end{tabular}\end{table*}

\par $\bullet$ $\textbf{SCUT-FBP}$ \cite{SCUT-FBP}: It is about facial beauty perception and contains 500 facial  images.  Each facial image is associated with an 5-D label distribution. The  256-D LBP, 256-D HOG and 224-D GIST features are extracted. 

\par $\bullet$ $\textbf{Emotion6}$: It has 1980  emotion images. Each emotion image is annotated with an 6-D label distribution. Three sets of features are extracted, i.e., 256-D LBP, 192-D RGB Hist and 192-D pixel average. 

\par $\bullet$ $\textbf{fbp5500}$ \cite{fbp5500}: It is a facial beauty perception dataset that includes 5500 facial  images. The label distribution is in 5 dimensions, and the 256-D LBP, 192-D HOG and 192-D Gabor features are extracted for each image.  
 
\par $\bullet$ $\textbf{RAF-ML}$ \cite{RAF_ML}: It is about facial expression recognition and has 4908  facial images. The facial image is associated with an 6-D label distribution. Three sets of features, i.e., 200-D LBP, 200-D baseDCNN and 200-D DBM${}_-$CNN, are employed to represent an image.

\par $\bullet$ $\textbf{Twitter-LDL }$ \cite{Twitter_and_Flickr}: It is an image visual sentiment dataset, and 2000  images are used.   Each image is depicted by an 8-D label distribution, and the 256-D LBP, 224-D HOG and 192-D RGB Hist features are extracted.

\par $\bullet$ $\textbf{Filckr-LDL}$ \cite{Twitter_and_Flickr}:  It is about  image visual sentiment, and 2000  images are utilized. Each image is annotated with an 8-D label distribution, and three sets of features are extracted, i.e., 256-D LBP, 224-D HOG and 192-D HSV Hist.  

\subsection{Baselines and Experimental Settings}
Considering that MVSS-LDL is the first study in multi-view LDL, we  compare MVSS-LDL with the single-view LDL baselines, which include the supervised single-view LDL approaches (i.e., LDL-LRR \cite{LDL-LRR}, LDL-LDM \cite{LDL-LDM} and LDL-SCL \cite{LDL-SCL}), and semi-supervised single-view LDL approaches (i.e. S2LDL \cite{SLDL-CO}, IncomLDL \cite{IncomLDL} and LDM-Incom  \cite{LDL-LDM}). \vspace{+0.2cm}

\par $\bullet$ \textbf{LDL-LRR} \cite{LDL-LRR}: It is a supervised single-view LDL approach that  formulates a  ranking function to extract the semantic patterns from label distributions.

\par $\bullet$ \textbf{LDL-SCL} \cite{LDL-SCL}: It is a supervised single-view LDL approach that explores the label relations on local samples and performs feature augmentation via the local relational vectors.

\par $\bullet$ \textbf{LDL-LDM} \cite{LDL-LDM}: It is a supervised single-view LDL approach that captures the global and local label dependencies through manifold learning of label distributions.  

\par $\bullet$ \textbf{LDM-Incom } \cite{LDL-LDM}: It is a semi-supervised single-view LDL approach. It is an extension of LDL-LDM to incomplete labeled data. By treating the unlabeled data as incomplete labeled data, it can be applied to  semi-supervised LDL.

\par $\bullet$ $\textbf{S2LDL}$ \cite{SLDL-CO}: It is a semi-supervised single-view LDL approach that adopts the co-regularization framework and learns the LDL model on labeled and unlabeled data. 

\par $\bullet$ \textbf{IncomLDL} \cite{IncomLDL}: It is a semi-supervised single-view LDL approach that uses the trace norm minimization technique to exploit label correlations.

\par All the baselines are single-view approaches and cannot be straightforwardly extended to multi-view scenarios. To adapt them to multi-view scenarios, multi-view features are fused into a concentrated vector, and the LDL models are trained using these  vectors.    Moreover, the parameter settings of MVSS-LDL and baselines are as follows. For LDL-LRR, the parameter $\lambda$ is chosen in $10^{[-6:1:1]}$, and  $\beta$ is picked up in $10^{[-3:1:3]}$. For LDL-SCL,  $\lambda_1$, $\lambda_2$ and $\lambda_3$ are picked up in $10^{[-1:1:3]}$. For LDL-LDM and LDM-Incom, the parameters $\lambda_1$, $\lambda_2$ and $\lambda_3$ are selected in $10^{[-3:1:3]}$, and  the parameter $g$ is chosen from 1 to 14. For  S2LDL, the hyper-parameters are chosen in $10^{[-5:1:5]}$. The parameters $\lambda$, $\beta$, $\alpha$ and $\gamma$ are fixed as $10^{-1}$, $10^{-3}$, $10^{-2}$ and  $10^{-3}$, respectively.    For IncomLDL, the parameter $\lambda$ is chosen in $2^{[-3:1:3]}$, and  $\gamma$ and $\epsilon$ are fixed as $2$ and $10^{-5}$, respectively.  For   MVSS-LDL, the number of nearest neighbors $k$ is fixed as 10,  and the regularized parameters $\gamma$, $\sigma$,  $\mu_1$ and  $\mu_2$ are  fixed as 100, 1000, 0.1 and 10, respectively.  The parameter  $\lambda$ is tuned in $10^{[-3:1:3]}$.

\begin{table*}[]
\renewcommand\arraystretch{1.4}
\centering
\caption{The $p$-values of Wilcoxon statistical test.}
\label{wilcoxon}
\begin{tabular}{ccccccc}
\hline
Compared Methods & Chebyshev & Clark    & Canberra & KL Divergence & Cosine   & Intersection \\ \hline
LDl-LRR	&	\textbf{win}[4.9e-04]	&	\textbf{win}[4.9e-04]	&	\textbf{win}[4.9e-03]	&	\textbf{win}[4.9e-04]	&	\textbf{win}[9.8e-04]	&	win[9.3e-03]	\\
LDL-SCL	&	\textbf{win}[1.2e-02]	&	\textbf{win}[4.9e-04]	&	\textbf{win}[9.8e-04]	&	\textbf{win}[1.5e-03]	&	\textbf{win}[4.9e-03]	&	\textbf{win}[4.2e-02]	\\
LDL-LDM	&	\textbf{win}[9.3e-03]	&	\textbf{win}[6.8e-03]	&	\textbf{win}[4.9e-04]	&	\textbf{win}[4.9e-04]	&	\textbf{win}[4.9e-04]	&	\textbf{win}[4.9e-04]	\\
LDM-Incom	&	\textbf{win}[9.8e-04]	&	\textbf{win}[4.9e-04]	&	\textbf{win}[9.8e-04]	&	\textbf{win}[1.5e-03]	&	\textbf{win}[4.9e-04]	&	\textbf{win}[4.9e-04]	\\
S2LDL	&	\textbf{win}[4.9e-04]	&	\textbf{tie}[9.1e-01]	&	\textbf{win}[2.7e-02]	&	\textbf{win}[4.9e-03]	&	\textbf{win}[9.3e-03]	&	\textbf{win}[1.2e-02]	\\
IncomLDL	&	\textbf{win}[4.9e-03]	&	\textbf{tie}[8.5e-01]	&	\textbf{win}[4.9e-04]	&	\textbf{win}[4.9e-04]	&	\textbf{win}[4.9e-04]	&	\textbf{win}[4.9e-04] \\
\hline
\end{tabular}
\end{table*}

\subsection{Experimental Result}
In this subsection, the  ratio of labeled samples is set to be 10\%. Only 10\% of the training samples are annotated, and the remaining 90\% are  unannotated.
 The performance of MVSS-LDL and baselines are assessed via ten-fold cross validation. The results under six evaluation metrics are illustrated in Table \ref{result}. The best results are emphasized in boldface.   An upward arrow (↑) represents that a higher value is preferable, and a downward arrow (↓) represents that a lower value is desirable. From the results in Table \ref{result}, we derive the following findings.

\par Firstly, MVSS-LDL  demonstrates markedly superior performance to the supervised single-view   LDL approaches (i.e., LDL-LRR, LDL-LDM and LDL-SCL) across all benchmark datasets.  Taking SCUT-FBP as an example, the Chebyshev value of MVSS-LDL is 0.3180, which is  lower than LDL-LRR (0.4640), LDL-LDM (0.7394) and LDL-SCL (0.7267) at 0.146, 0.4214 and 0.4087, respectively.  Additionally, on the Emotion6 dataset, the Canberra value of MVSS-LDL is 3.7421, which is lower than LDL-LRR (4.7213), LDL-LDM (4.1300) and LDL-SCL (4.4403) at 0.9792, 0.3879 and 0.6982, respectively.   LDL-LRR, LDL-LDM and  LDL-SCL  are supervised  LDL approaches, which use  the labeled data to construct LDL models. However, when there is only a small amount of labeled data available during the training process, the learned classifier may be prone to overfitting.  Different from the supervised LDL approaches,  MVSS-LDL is a semi-supervised LDL  approach, which  leverages both labeled and unlabeled data to enhance the LDL model. By integrating the unlabeled data, the classifier is refined to be more accurate even if only  a small amount of labeled data is accessible.

 Secondly, MVSS-LDL  achieves notably better results compared to the  semi-supervised single-view LDL approaches (i.e.,  S2LDL, IncomLDL and LDM-Incom).  On the one hand,  MVSS-LDL outperforms S2LDL across all benchmark datasets. Taking the Twitter-LDL  dataset as an example, the Cosine values of MVSS-LDL and S2LDL are 0.5922 and 0.4096, respectively. MVSS-LDL obtains an improvement at 0.1826. 
 Moreover, the Intersection values of MVSS-LDL and S2LDL are 0.3265 and 0.2107, respectively.  MVSS-LDL reports 0.1158 improvements relative to S2LDL.   On the other hand, MVSS-LDL shows improved performance over IncomLDL in 34 out of 36 configurations (6 evaluation metrics and 6 datasets). Except for the KL Divergence of  SCUT-FBP and Clark of Twitter-LDL, MVSS-LDL attains explicitly better performance than IncomLDL. Furthermore, MVSS-LDL surpasses LDM-Incom in 34 out of 36 configurations. Except for the KL Divergence of Twitter-LDL and Flickr-LDL,  MVSS-LDL delivers more desirable classification results than LDM-Incom.  Taking RAF-ML as an example, the Cosine value of MVSS-LDL is 0.6848 that is  higher than IncomLDL (0.5755) and LDM-Incom (0.6031) at 0.1093 and 0.0817, respectively.  The superior performance of MVSS-LDL over S2LDL, IncomLDL and LDM-Incom  demonstrates the efficacy of multi-view learning.
 S2LDL, IncomLDL and LDM-Incom are  single-view LDL algorithms, which handles the data with only one view. In order to adapt them to multi-view data,  the multi-view features are fused to form a concatenated vector, and the concatenated vectors are then used to train the LDL model.   Nevertheless, combining a number of views into a concatenated vector may ignore the correlation  between  views, which limits their discriminative ability. In contrast with S2LDL, IncomLDL and LDM-Incom,  MVSS-LDL is a multi-view LDL algorithm, which can integrate the consistency information and complementarity information of distinct views into enhancing the LDL model.

Lastly, the Wilcoxon signed-rank test is conducted to  investigate whether the performance improvement of MVSS-LDL is significant compared to the baselines. As a non-parametric pairwise test, it calculates statistical significance by comparing MVSS-LDL's classification outcomes with those of the baselines. The null hypothesis is that it has no difference in the distribution of classification outcomes. When the  $p$-value falls below the threshold 0.05, we reject the null hypothesis and conclude that MVSS-LDL demonstrates statistically significant improvement. The statistical comparisons of MVSS-LDL and baselines with the Wilcoxon signed-rank tests are summarized in Table \ref{wilcoxon}.
It can be seen that  MVSS-LDL demonstrates significantly better classification performance than the supervised single-view LDL approaches  (i.e., LDL-LRR, LDL-SCL and LDL-LDM ) and the semi-supervised single-view LDL approach (i.e., LDM-Incom) on all evaluation metrics.  Furthermore, MVSS-LDL significantly outperforms the  semi-supervised single-view LDL approaches (i.e.,  S2LDL and IncomLDL) on 5 out of 6 metrics.   These findings further verify the efficacy of multi-view learning on LDL tasks.

\subsection{Performance with Different Number of Labeled Samples}

\begin{figure*}[htbp]	
    \centering
    \subfloat[SCUT-FBP]{
        \includegraphics[width=0.312\linewidth]{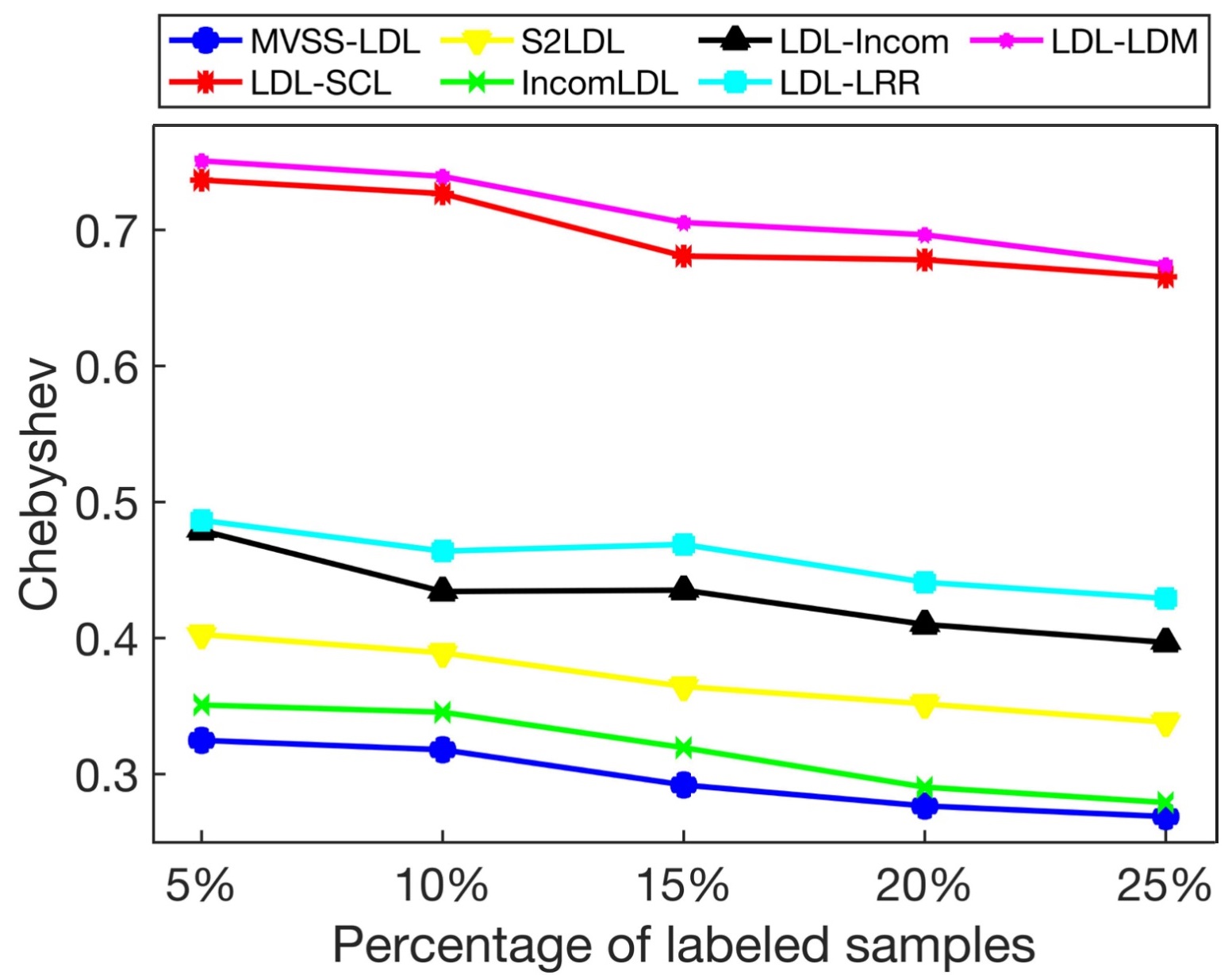}
        } \hfill
    \subfloat[Emotion6]{
        \includegraphics[width=0.312\linewidth]{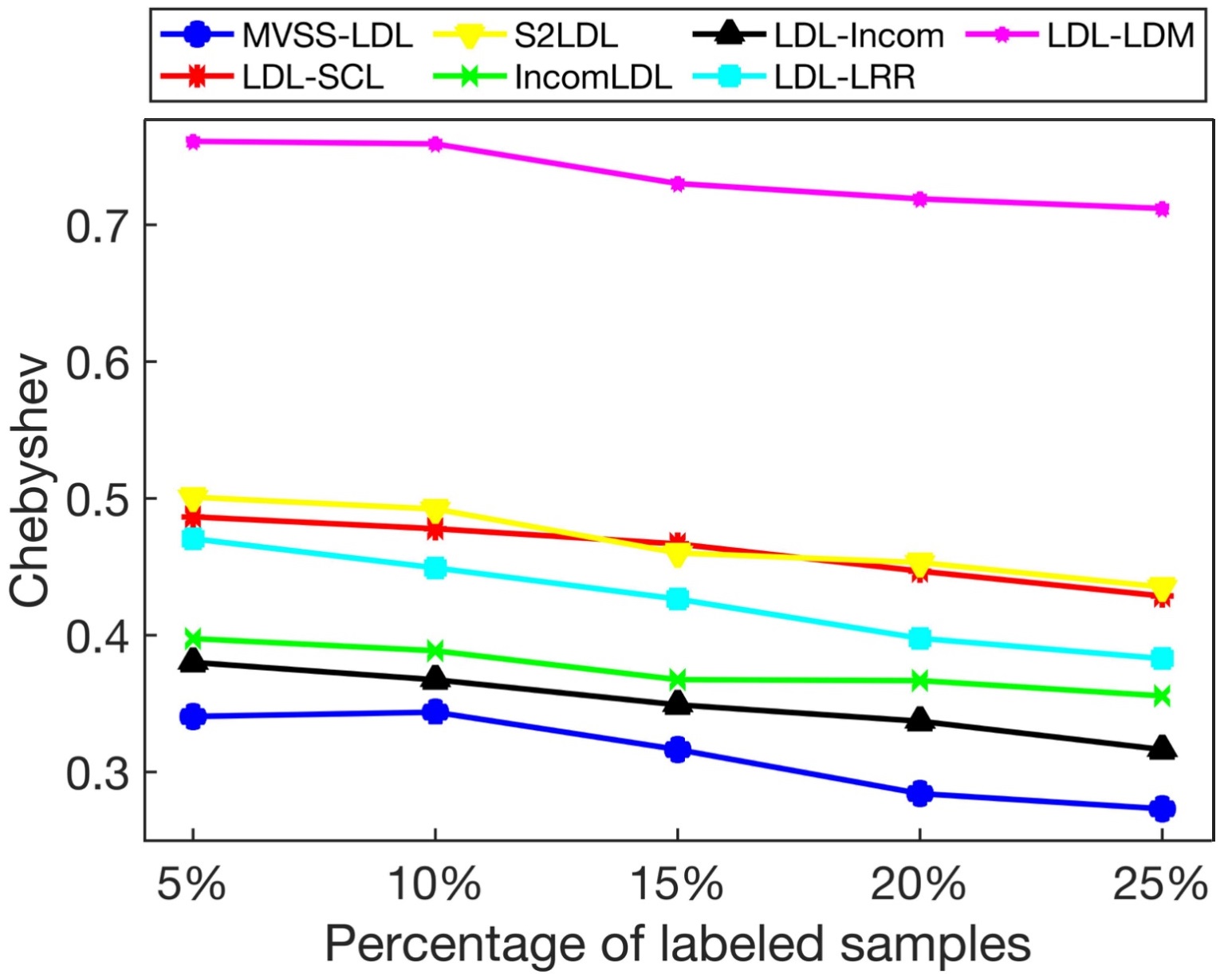}
        } \hfill
        \subfloat[fbp5500]{
        \includegraphics[width=0.312\linewidth]{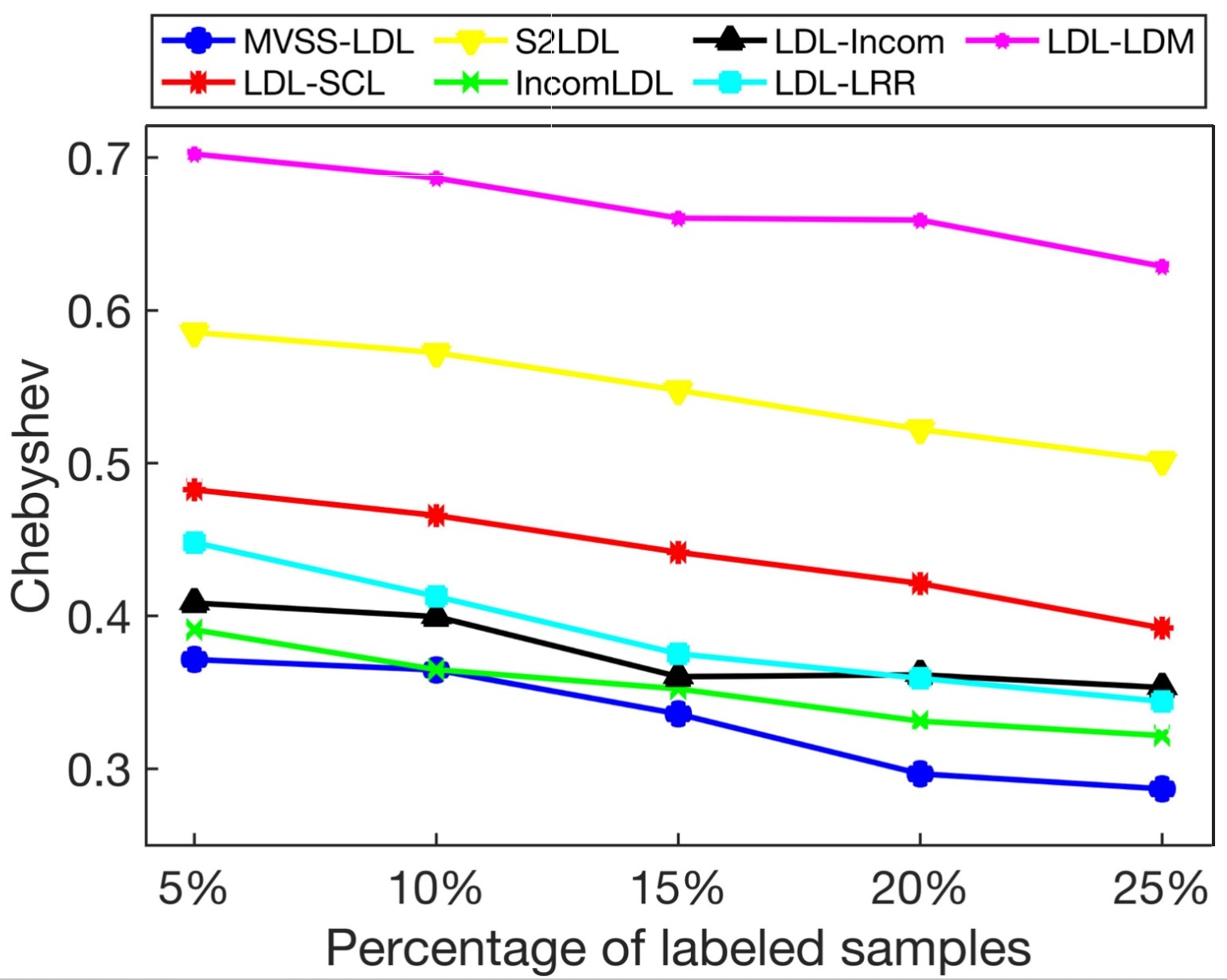}
        } \\
       \subfloat[RAF-ML] {
        \includegraphics[width=0.312\linewidth]{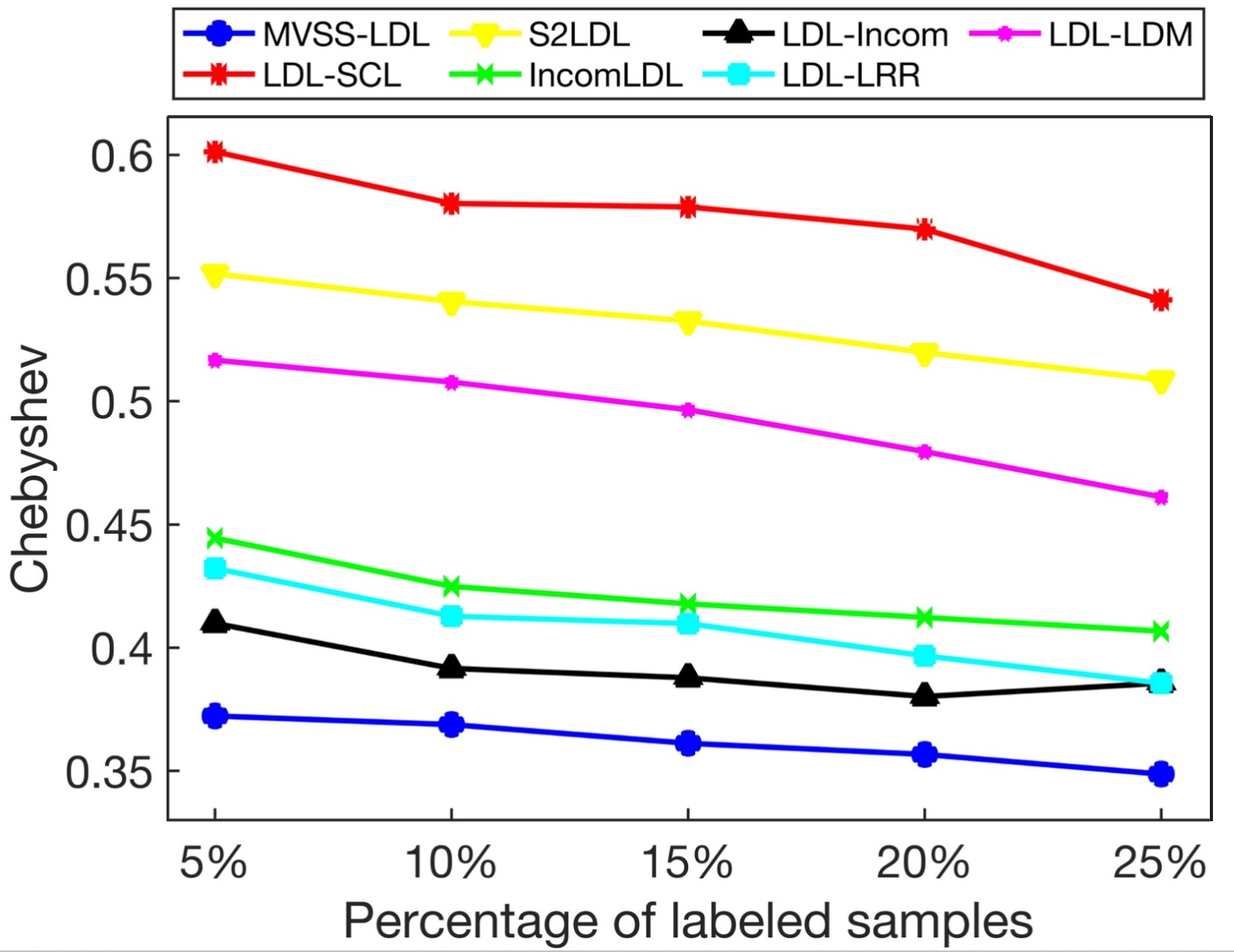}
        }\hfill
    \subfloat[Twitter-LDL]{
        \includegraphics[width=0.312\linewidth]{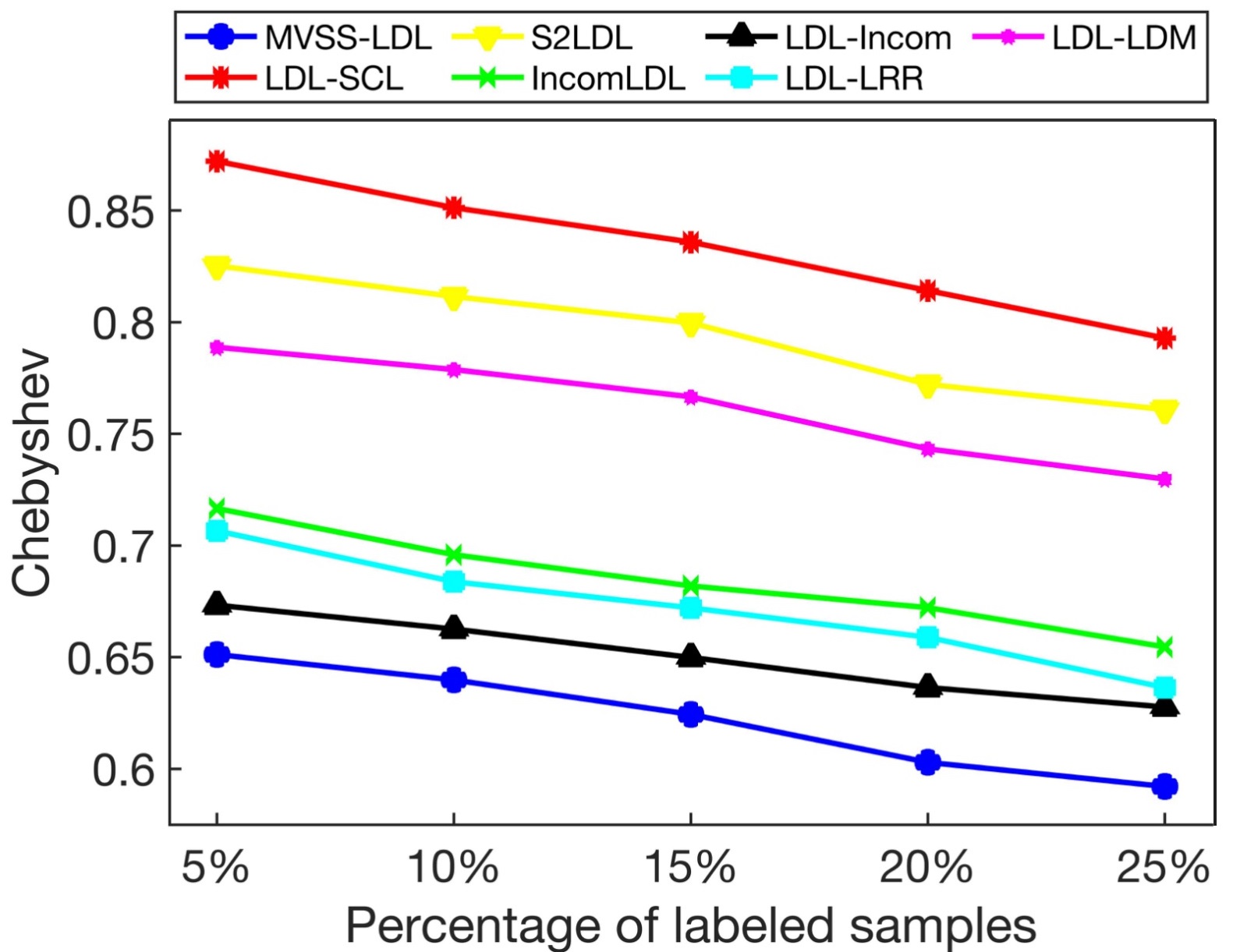}
        }\hfill
        \subfloat[Flickr-LDL]{
        \includegraphics[width=0.312\linewidth]{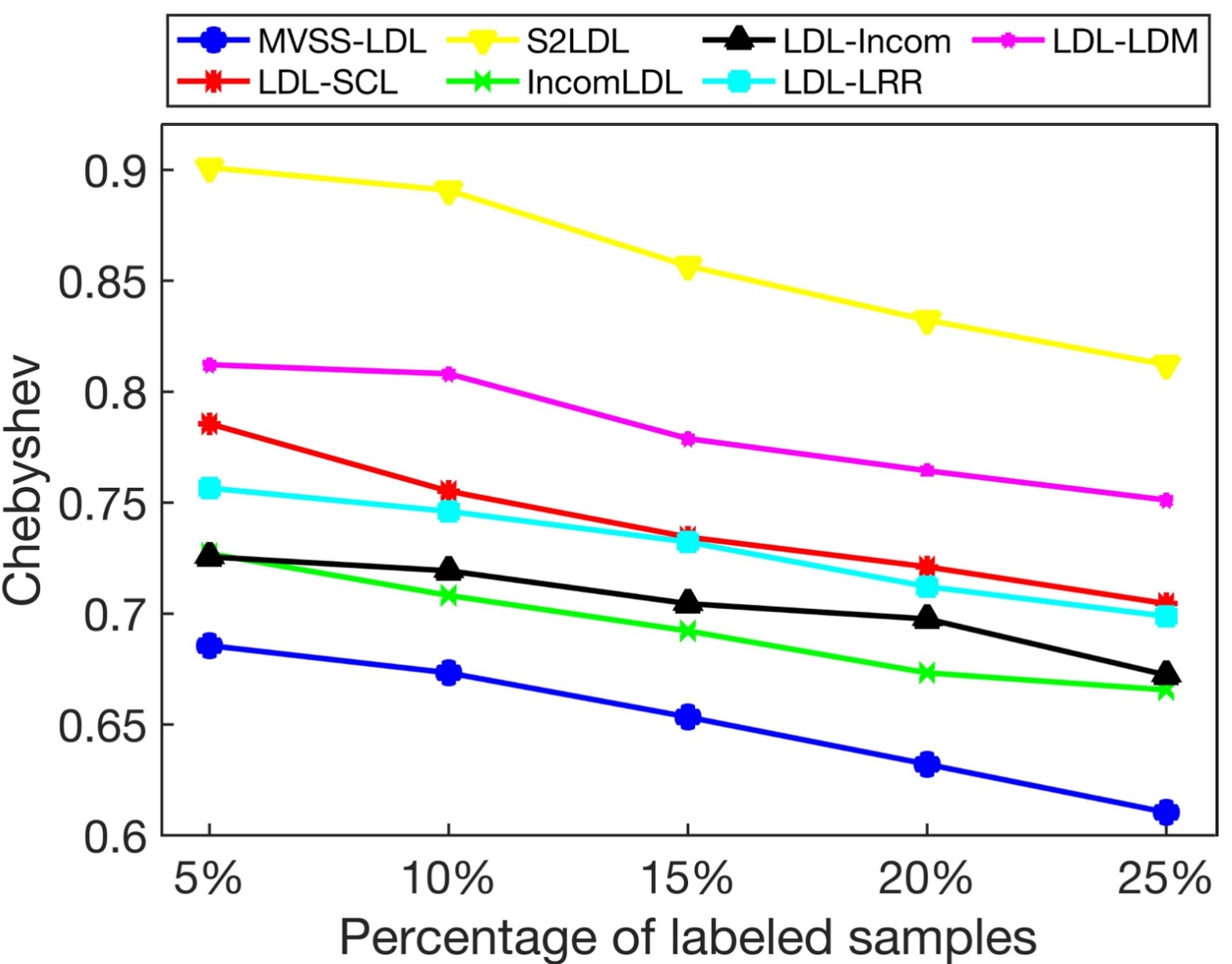}
        }             \caption{The Chebyshev value ($\downarrow$)  of MVSS-LDL and the baselines with different number of labeled samples.}
        \label{different label}
\end{figure*}

\par To evaluate the performance of MVSS-LDL with different number of labeled training samples, we set the percentage of labeled  samples to be $5\%$, $10\%$, $15\%$, $20\%$ and $25\%$, respectively, and the corresponding Chebyshev values are reported, as illustrated  in Fig. \ref{different label}. In Fig. \ref{different label}, the $x$-axis is the percentage of labeled  samples, and the  $y$-axis is the Chebyshev value. From   Fig. \ref{different label}, we have the following observations. In  one aspect, as   the percentage of labeled  samples increases, the performance of all approaches improves. This is because when more labeled samples are included in the training process, more classification information is attained, and the learned LDL model  becomes more precise. In the other aspect, MVSS-LDL consistently outperforms the baselines  on all datasets. Taking RAF-ML as an example, MVSS-LDL
achieves the lowest Chebyshev value as the percentage of labeled  samples goes up from $5\%$ to $25\%$. Different from the baselines that utilize only the single-view data, MVSS-LDL  constructs the  classifier on multi-view data, and incorporates the consistency information and complementarity information of multiple views into refining the classifier, which makes  our approach  achieve satisfactory classification performance even if only a small amount of  labeled data is accessible.

\subsection{Parameter Sensitivity Analysis}

\begin{figure*}[hbtp]	
    \centering
    \subfloat[SCUT-FBP]{
        \includegraphics[width=0.30\linewidth]{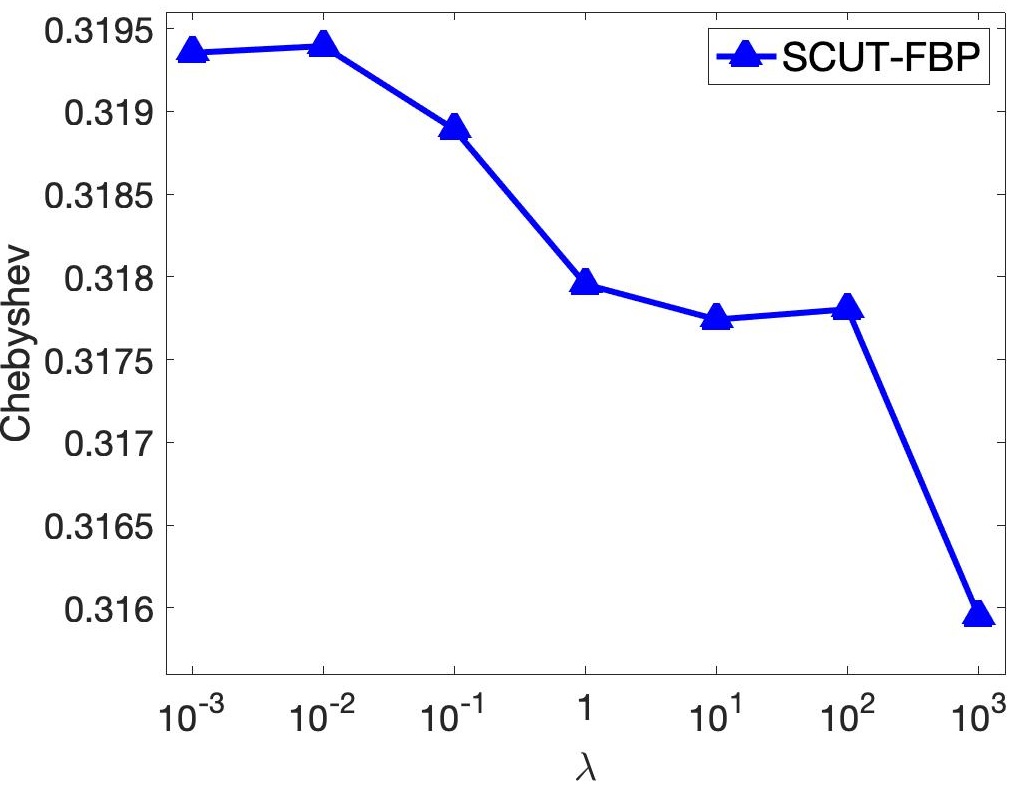}
        \label{SCUT-FBP_chebyshev}
        } \hfill
    \subfloat[Emotion6]{
        \includegraphics[width=0.30\linewidth]{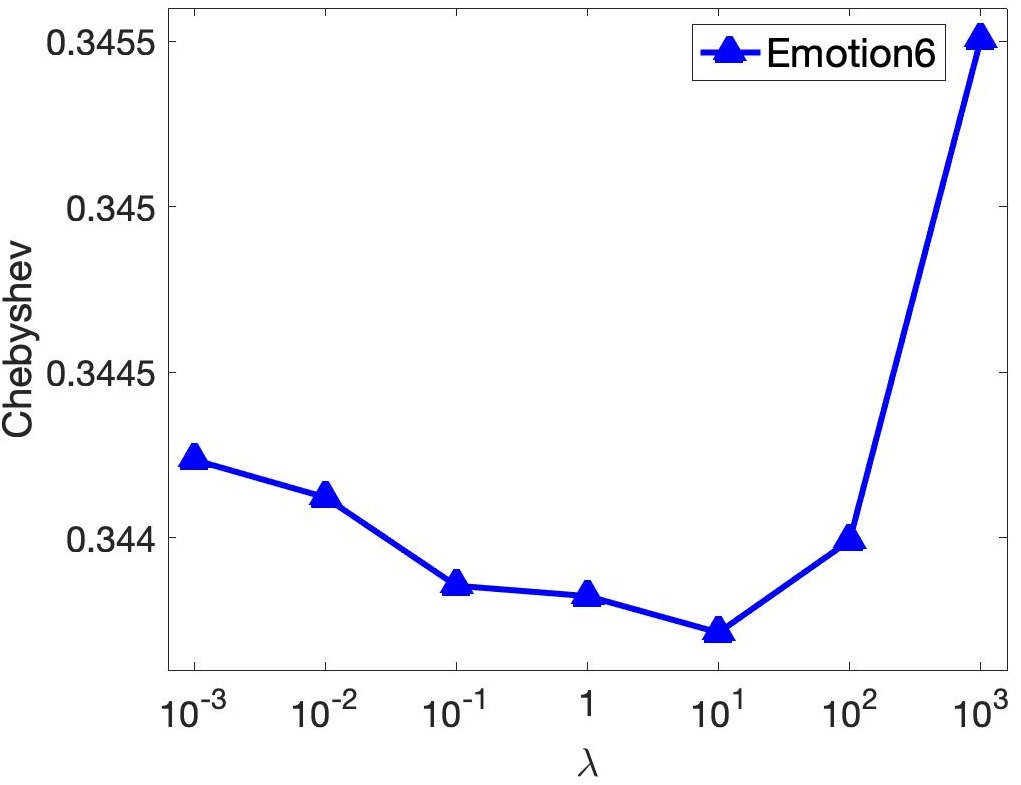}
        } \hfill
        \subfloat[fbp5500]{
        \includegraphics[width=0.30\linewidth]{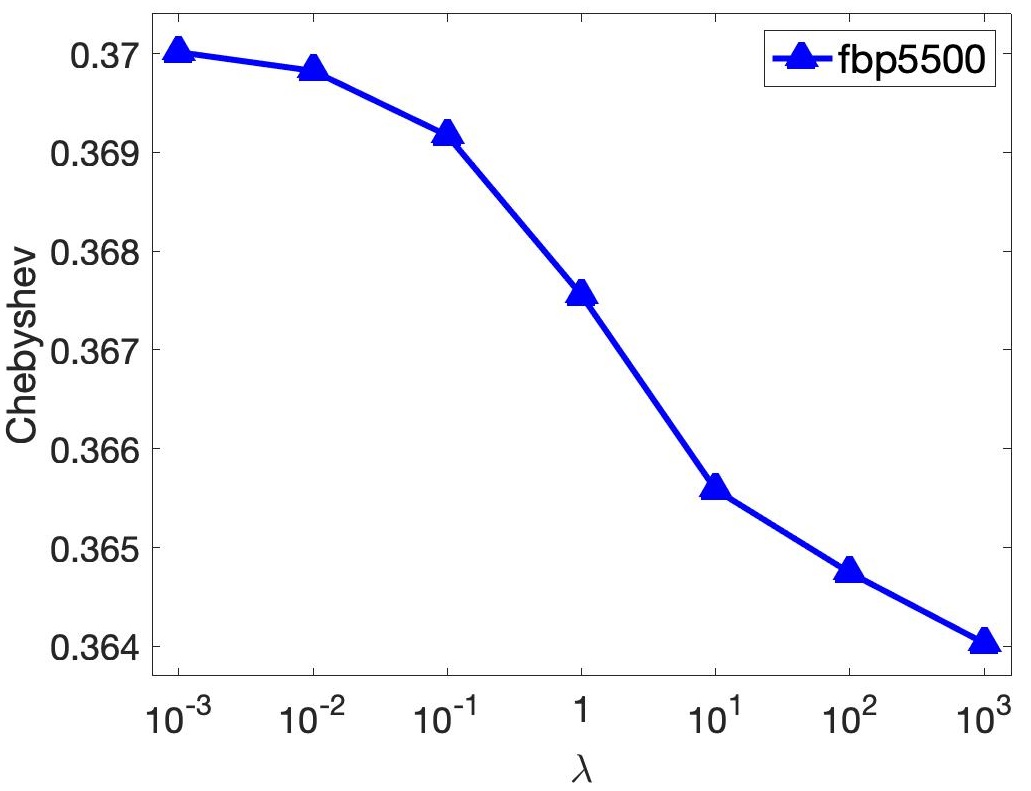}
        } \\
       \subfloat[RAF-ML] {
        \includegraphics[width=0.3\linewidth]{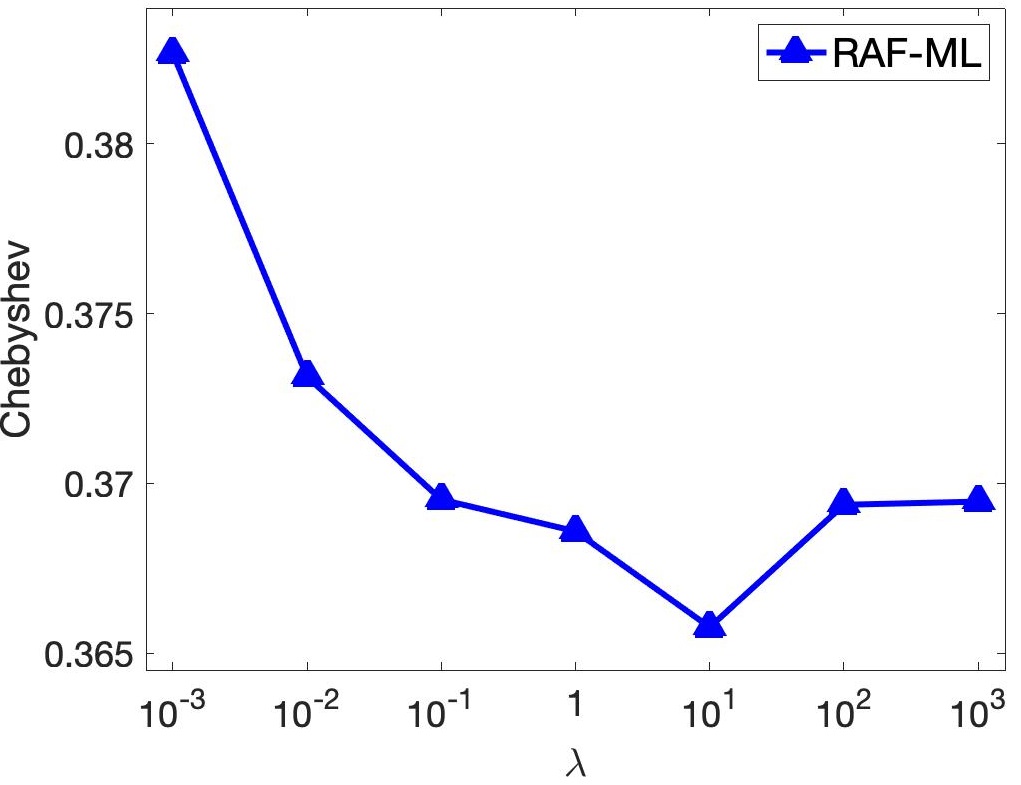}
        \label{RAF-ML_cosine}
        }\hfill
    \subfloat[Twitter-LDL]{
        \includegraphics[width=0.3\linewidth]{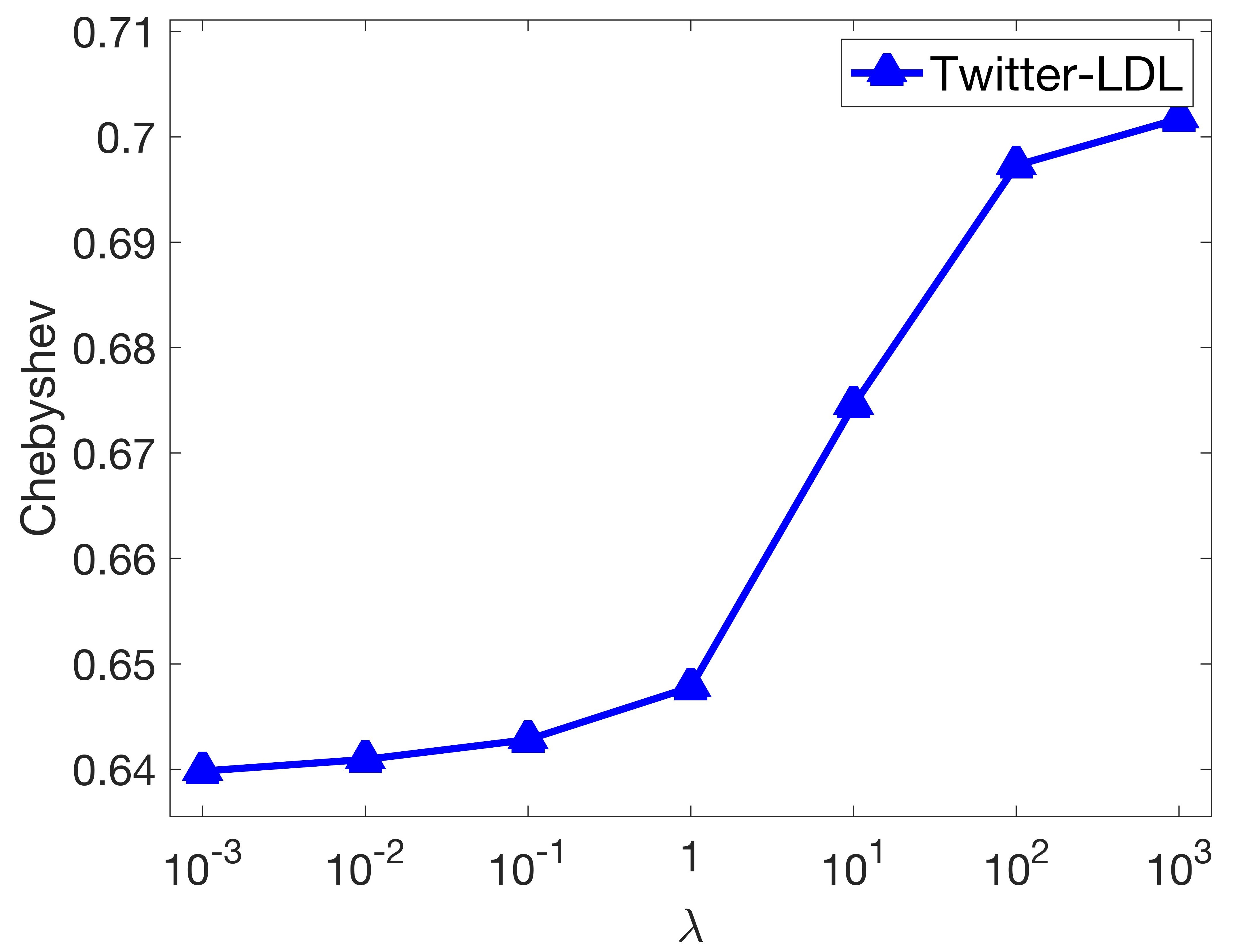}
        }\hfill
        \subfloat[Flickr-LDL]{
        \includegraphics[width=0.3\linewidth]{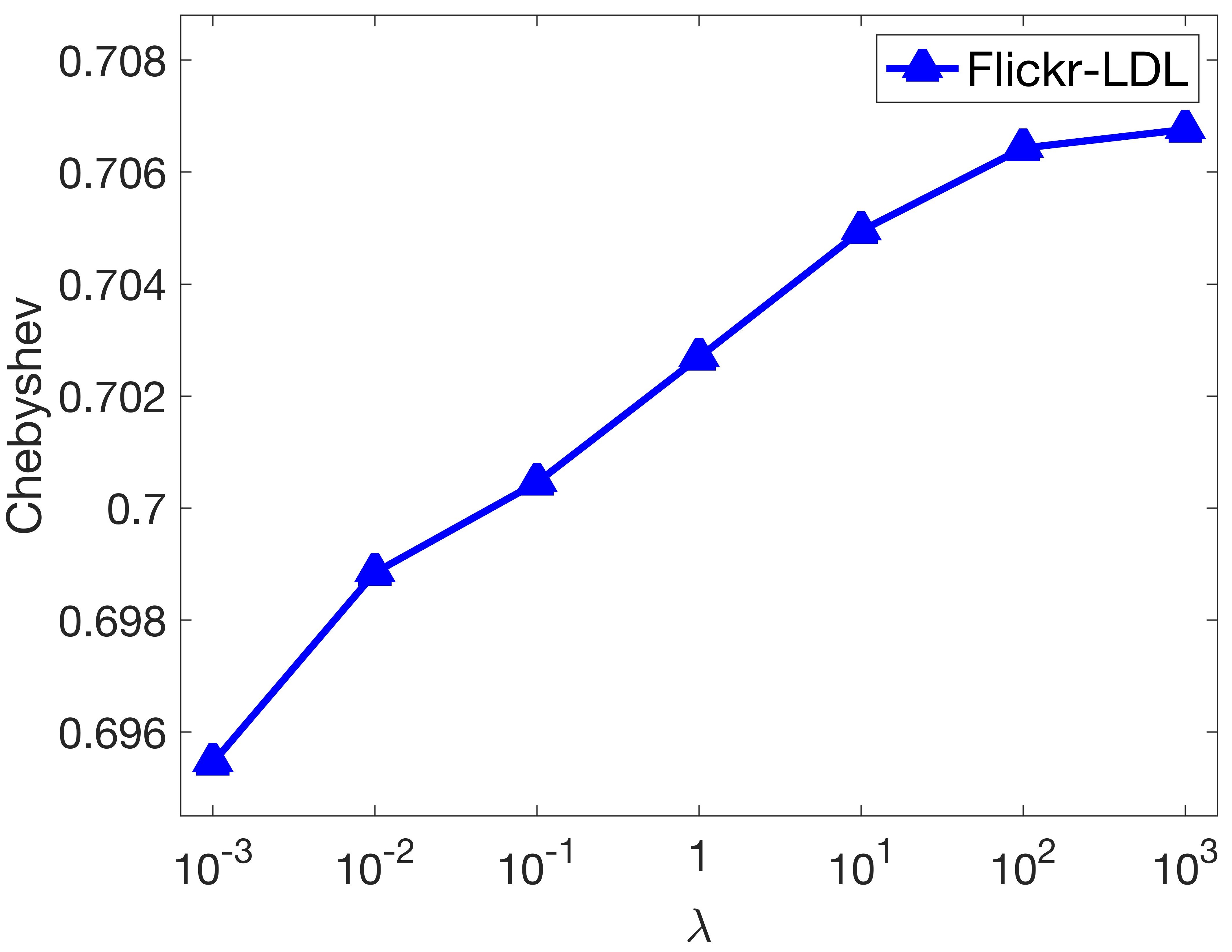}}
                    \caption{The Chebyshev value  ($\downarrow$)  of MVSS-LDL with different value of parameter $\lambda$.}
        \label{fig:sensitivity}
\end{figure*}

 In the experiment, we only need to tune  the parameter $\lambda$, since  the number of nearest neighbors $k$ is fixed as 10, and the regularized parameters  $\gamma$,  $\sigma$,  $\mu_1$ and  $\mu_2$ are  fixed as  100, 1000, 0.1 and 10, respectively. Thus,  the impact of parameter $\lambda$ on MVSS-LDL is  investigated.  Fig. \ref{fig:sensitivity} illustrates the Chebyshev value of MVSS-LD when    $\lambda$ varies from $10^{-3}$ to $10^{3}$. 
 
In one aspect,  on the  SCUT-FBP, Emotion6, fbp5500  and RAF-ML  datasets, the lowest Chebyshev value  is obtained when $\lambda$ is relatively large (e.g., 10 or 1000). In the other aspect, on the Flickr-LDL  and Twitter-LDL  datasets, the lowest Chebyshev value  is attained when  $\lambda$ is relatively small (e.g., 0.001). This is because  $\lambda$ is associated with the term $ \sum \|(\mathbf{W}^{v})^{\top}\boldsymbol{x}_{i}^{v}- \boldsymbol{d}_{i}^v \|_{2}^{2}$.  For the labeled data, this term represents the difference of the  predicted label distributions  and ground-truth label distributions.  As shown in Table \ref{result}, the Chebyshev values of MVSS-LDL on the SCUT-FBP, Emotion6, fbp5500  and RAF-ML  datasets are  between 0.3180 and  0.3688, which are relatively small. This indicates that the  SCUT-FBP, Emotion6, fbp5500  and RAF-ML datasets are relatively easy to be classified, and   the difference between  the  predicted label distributions  and ground-truth label distributions is small, which enables the  value of $\lambda$ to be large.  Furthermore, the  Chebyshev values  of MVSS-LDL  on the Twitter-LDL  and Flickr-LDL  datasets are 0.6398 and 0.6733, receptively, which are relatively large. This implies that the Twitter-LDL  and Flickr-LDL  datasets are relatively difficult to be classified, and   the difference between the predicted label distributions and ground-truth label distributions is large. To balance it, a small  $\lambda$ value  is required.

 \subsection{Ablation Study}

\par
\begin{figure*}[htbp]	
    \centering
    \subfloat[Chebyshev$\downarrow$]{
        \includegraphics[width=0.312\linewidth]{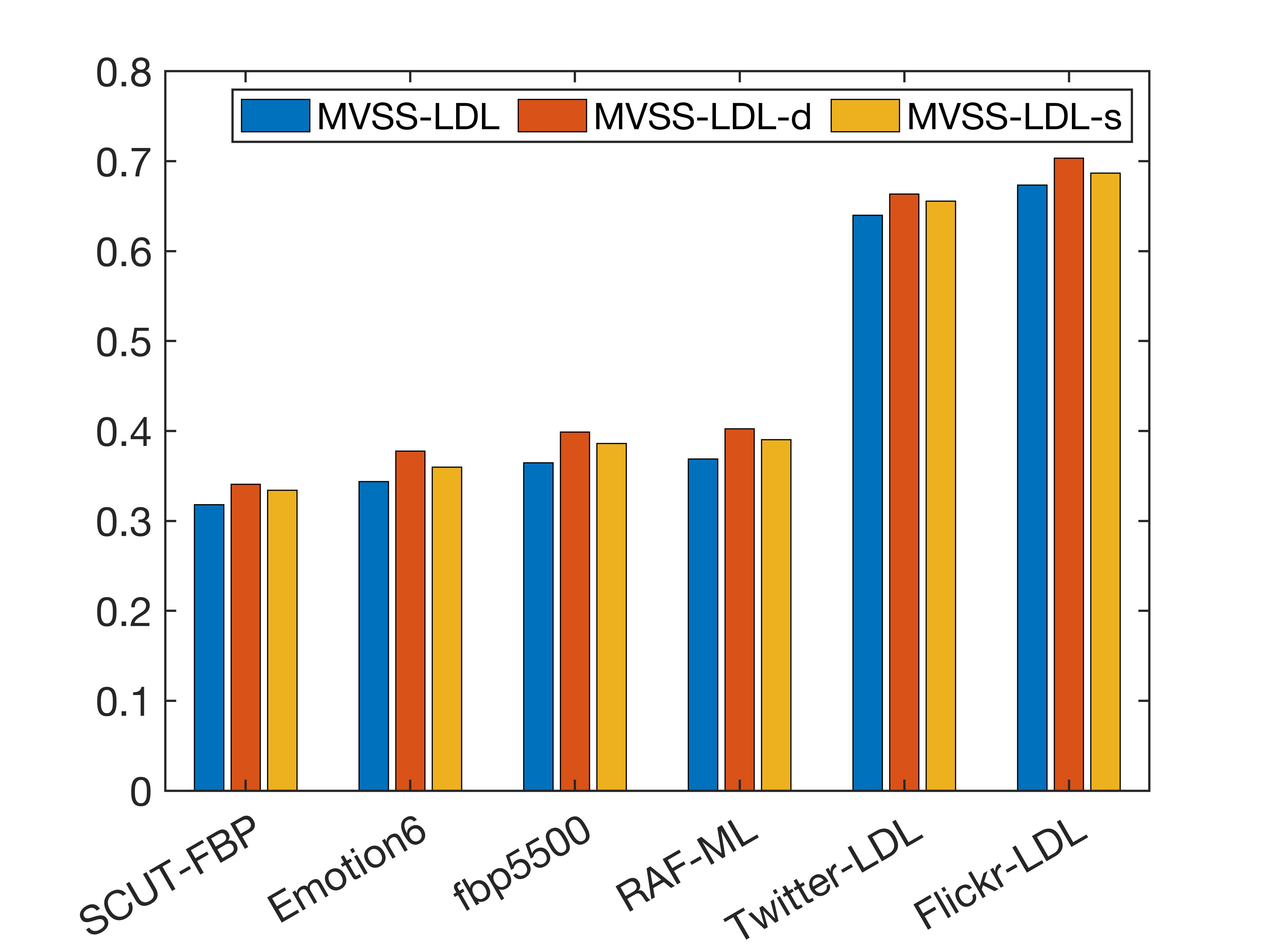}
        } \hfill
    \subfloat[Clark$\downarrow$]{
        \includegraphics[width=0.312\linewidth]{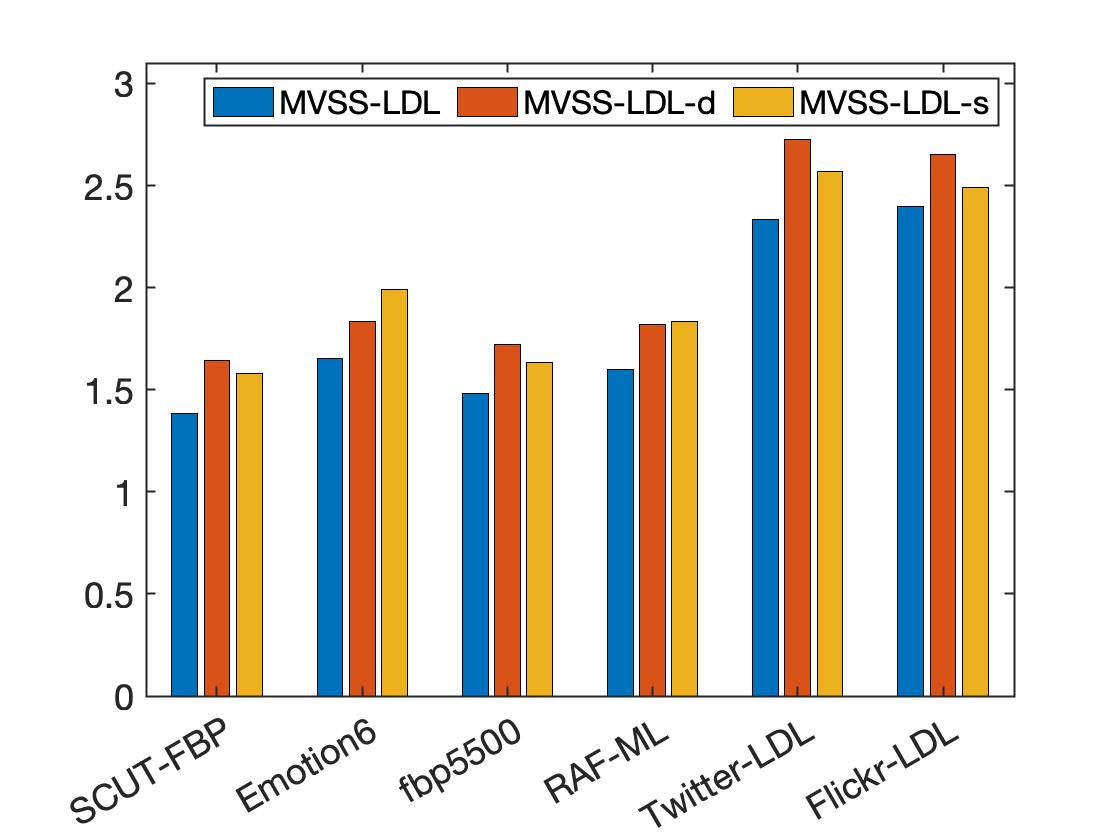}
        }\hfill
        \subfloat[Canberra$\downarrow$]{
        \includegraphics[width=0.312\linewidth]{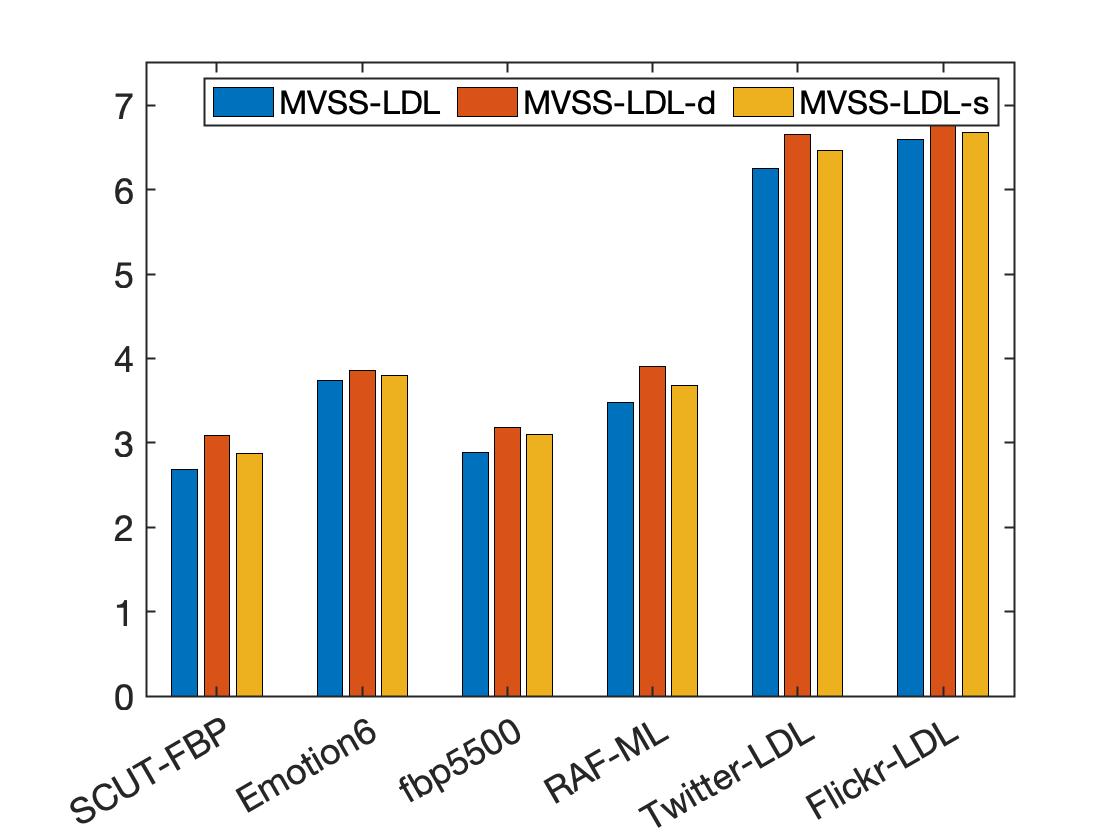}
        } \\
         \subfloat[KL divergence$\downarrow$]{
       \includegraphics[width=0.312\linewidth]{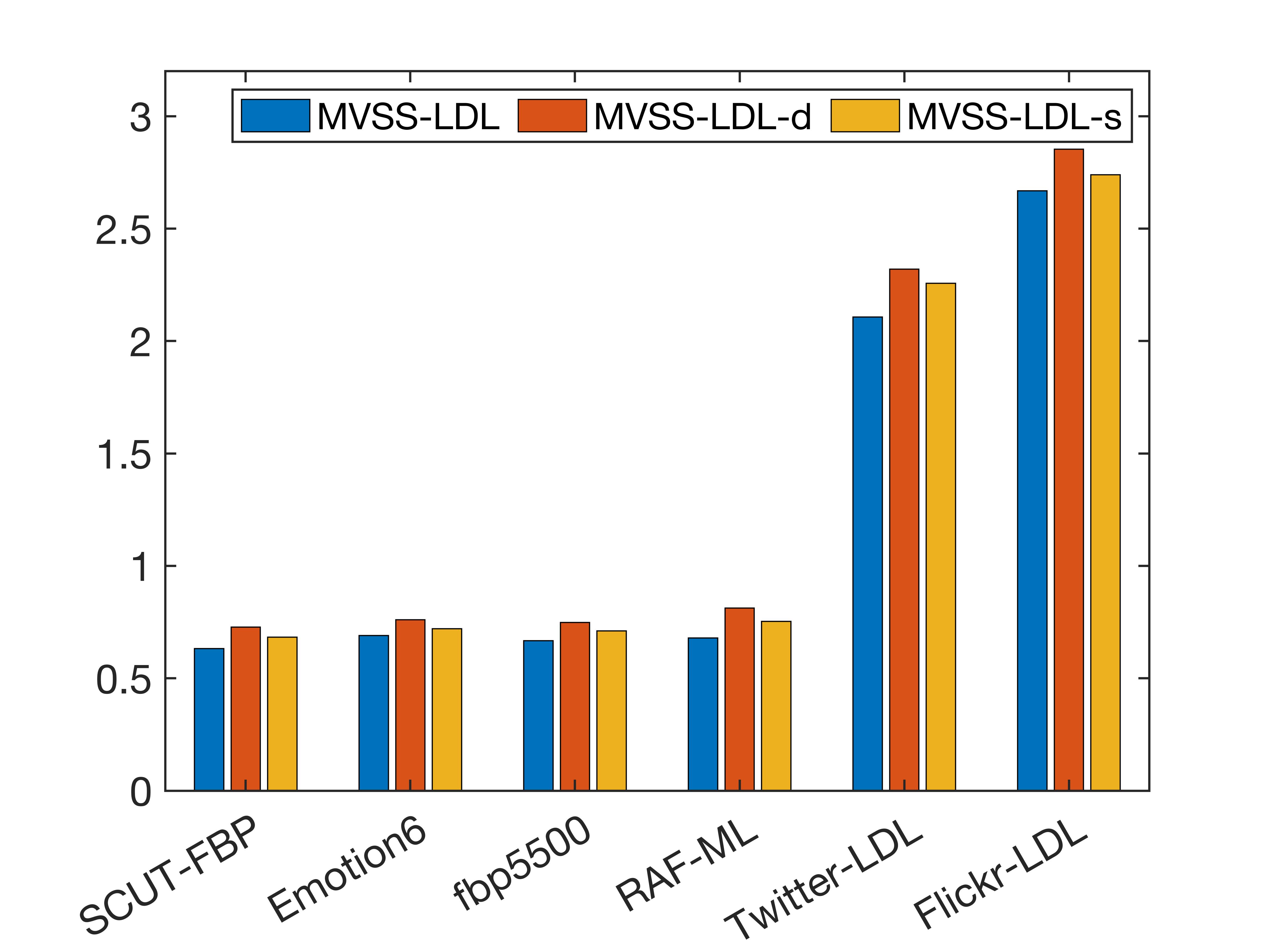}
      } \hfill
    \subfloat[Cosine$\uparrow$]{
        \includegraphics[width=0.312\linewidth]{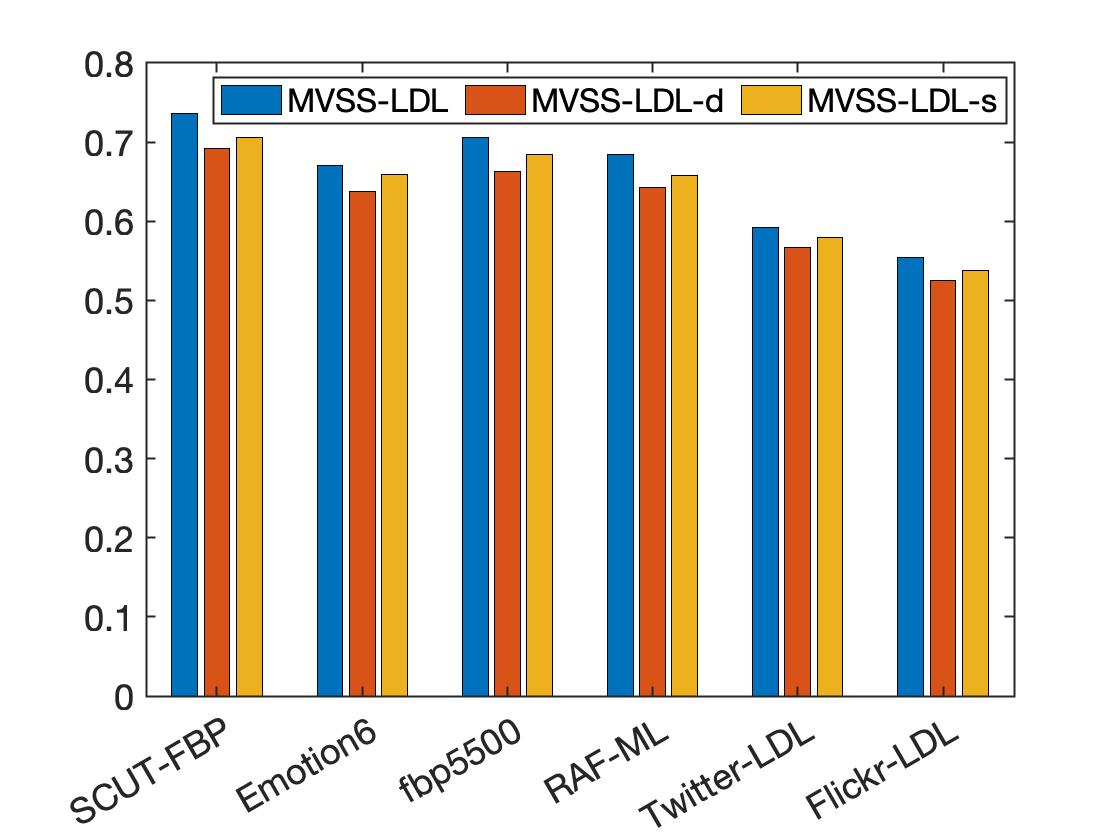}
        }\hfill
        \subfloat[Intersection$\uparrow$]{
        \includegraphics[width=0.312\linewidth]{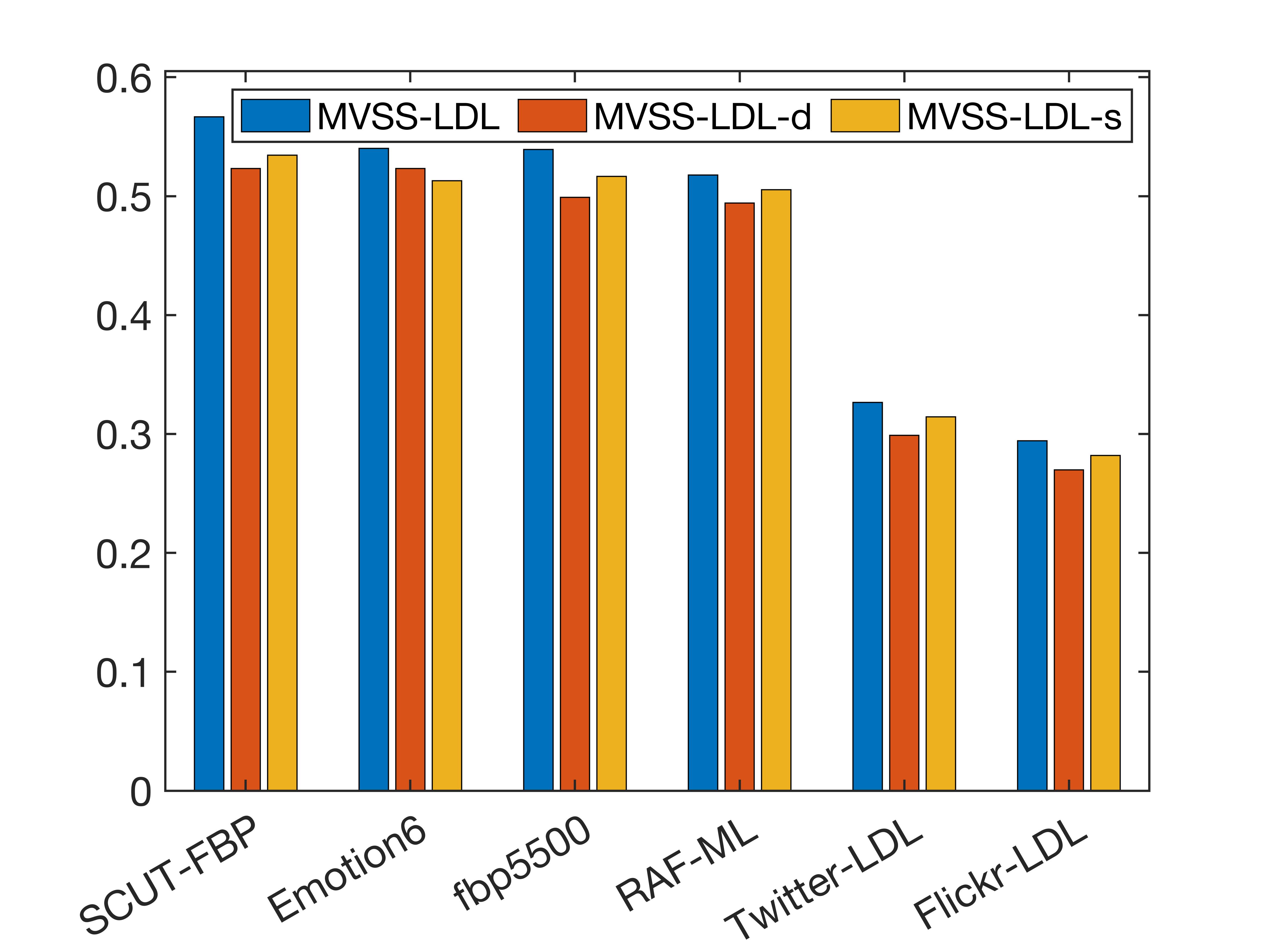}
        }
        \caption{The results of ablation experiments under all evaluation metrics.}
        \label{ablation}
\end{figure*}

 In order to verify the impact of the similarity consistency regularizer  and label distribution consistency regularizer on our approach, we form two  MVSS-LDL variants, i.e.,  MVSS-LDL-s and MVSS-LDL-d. In MVSS-LDL-s, the parameter $\sigma$ is set to be 0,  and the similarity consistency regularizer has no influence on the learning model. 
By comparing the performance of MVSS-LDL-s and MVSS-LDL, we can evaluate the  promotion  of the similarity consistency regularizer to the learning model. Similarly, in  MVSS-LDL-d, the parameter $\gamma$ is set to be 0, and  the label distribution consistency  regularizer plays little role in the learning the model. Fig. \ref{ablation} illustrates the results of MVSS-LDL, MVSS-LDL-s and MVSS-LDL-d on the experimental datasets.  In Fig. \ref{ablation}, it is seen that  MVSS-LDL achieves better classification performance than MVSS-LDL-s and MVSS-LDL-d in terms of all the evaluation metrics, which indicates the effectiveness of the similarity consistency regularizer and label distribution consistency  regularizer. The similarity consistency regularizer requires that a sample should have the same similarity to its $k$-nearest neighbors in different views. The label distribution consistency regularizer requires that different views of the same sample should have the same predicted label distributions. By incorporating these  regularizers, the consistency information of multiple views can be integrated into refining the LDL model, and the classification performance is improved.

\section{Conclusion}
In this paper, we put forward a multi-view semi-supervised LDL approach  MVSS-LDL, which  emphasize the complementarity of   local nearest neighbor structures   in multiple views. As far as we known, this is the first attempt on multi-view LDL. In one aspect,  MVSS-LDL explores  the local nearest neighbor structure of each view and complements the local nearest neighbor structure  of one view by incorporating the nearest neighbor information of other views. In the other aspect, a graph-based multi-view semi-supervised LDL model is constructed.  Substantial experiments have demonstrated 
that MVSS-LDL attains significantly better classification performance than the single-view LDL approaches. In the future, we would like to extend MVSS-LDL to  the online setting.




\end{document}